\documentclass{article}
\pdfoutput=1

\usepackage[nonatbib,preprint]{neurips_2023}

\usepackage[utf8]{inputenc} 
\usepackage[T1]{fontenc}    
\usepackage{hyperref}       
\usepackage{url}            
\usepackage{booktabs}       
\usepackage{amsfonts}       
\usepackage{nicefrac}       
\usepackage{microtype}      
\usepackage{xcolor}         
\usepackage{algorithm}      
\usepackage{blindtext}
\usepackage{graphicx}
\usepackage{multirow}
\usepackage{enumitem}

\title{A Comprehensive Evaluation of GPT-4V on Knowledge-Intensive Visual Question Answering}

\author{
Yunxin Li\thanks{ equal contributions, $^\dagger$ corresponding author},~~~~ Longyue Wang$^{*}$,~~~~ Baotian Hu$^{\dagger}$,~~~~ Xinyu Chen,~~~~ Wanqi Zhong, \\ \bf Chenyang Lyu,~~~~~ Wei Wang,~~~~~ Min Zhang \\
School of Computer Science and Technology, Harbin Institute of Technology, Shenzhen\\
\textit{liyunxin987@163.com, vincentwang0229@gmail.com, hubaotian@hit.edu.cn}\\
}

\begin{document}

\maketitle

\begin{figure}[h]
    \centering
    \includegraphics[width=0.89\textwidth]{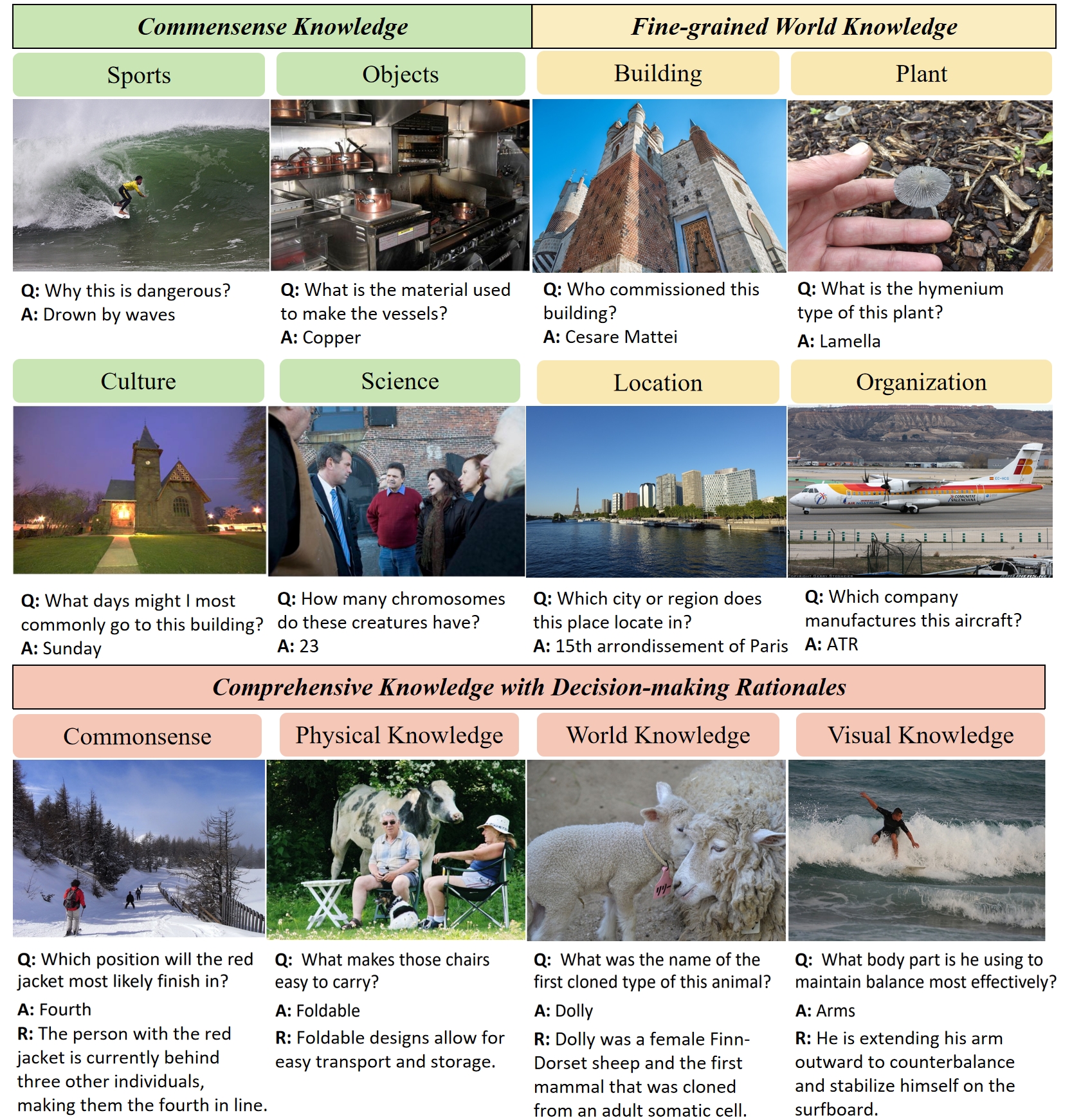}
    \caption{The illustration of our evaluation framework for knowledge-intensive visual question answering (VQA) tasks, which respectively focus on different dimensions of knowledge: {\em commonsense knowledge}, {\em fine-grained world knowledge}, and {\em comprehensive knowledge with decision-making rationales}. We aim to cover a wide spectrum of knowledge types and categories, ensuring a holistic evaluation of GPT-4V's capabilities of understanding, reasoning, and interpreting.}
    \label{fig:case_intro}
\end{figure}

\begin{abstract}

The emergence of multimodal large models (MLMs) has significantly advanced the field of visual understanding, offering remarkable capabilities in the realm of visual question answering (VQA). Yet, the true challenge lies in the domain of knowledge-intensive VQA tasks, which necessitate not just recognition of visual elements, but also a deep comprehension of the visual information in conjunction with a vast repository of learned knowledge. To uncover such capabilities of MLMs, particularly the newly introduced GPT-4V and Gemini, we provide an in-depth evaluation from three perspectives: 1) \textit{Commonsense Knowledge}, which assesses how well models can understand visual cues and connect to general knowledge; 2) \textit{Fine-grained World Knowledge}, which tests the model's skill in reasoning out specific knowledge from images, showcasing their proficiency across various specialized fields; 3) \textit{Comprehensive Knowledge with Decision-making Rationales}, which examines model's capability to provide logical explanations for its inference, facilitating a deeper analysis from the interpretability perspective. Additionally, we utilize a visual knowledge-enhanced training strategy and multimodal retrieval-augmented generation approach to enhance MLMs, highlighting the future need for advancements in this research direction. Extensive experiments indicate that: a) GPT-4V demonstrates enhanced explanation generation when using composite images as few-shots; b) GPT-4V and other MLMs produce severe hallucinations when dealing with world knowledge; c) Visual knowledge enhanced training and prompting technicals present potential to improve performance.

\end{abstract}

\section{Introduction}

Visual Question Answering (VQA) tasks serve as a benchmark for evaluating an AI system's ability to interpret and reason about visual content in conjunction with textual information, holding significant importance in both academic research and practical applications. By addressing the complexities and interdisciplinary challenges posed by VQA, researchers and practitioners can push forward the boundaries of what's possible in AI, making strides toward machines that can see, understand, and interact with the world in a human-like manner. Recently proposed Multimodal Large Models (MLMs)~\cite{li2023blip2,liu2023improved, liu2023visual,zhang2023llama} achieve significant success in visual-language understanding and reasoning, especially for Visual Question Answering (VQA) scenario centered around image understanding. To our knowledge, when humans interact with multi-modal artificial intelligence systems, they often expect to acquire valuable information that they do not know and introduce information-seeking questions, which involves knowledge-intensive challenges.
This raises a pertinent inquiry: \textit{how do MLMs-based VQA systems fare against knowledge-intensive information-seeking questions?} 
It is thus imperative to investigate MLMs' performance in the knowledge-intensive VQA scenario, as this will not only gauge their visual reasoning ability based on their stored knowledge but also pave the way for augmenting their capabilities in visual question answering and improving their credibility.

In this study, we concentrate on evaluating the capabilities of advanced multimodal large models, with a specific emphasis on GPT-4V(ision), in the context of knowledge-intensive VQA tasks. We have designed a comprehensive benchmark by reconfiguring existing knowledge-based VQA datasets, which will allow us to probe the reasoning ability of these models on common sense and fine-grained world knowledge, and their capacity to generate rationales for logical decision-making. The overall evaluation mainly consists of: 1) \textit{Performance on Wide Knowledge Scope}, where we design a knowledge-intensive VQA benchmark, including commonsense and world knowledge with more than ten knowledge categories, to assess advanced MLMs (including GPT-4V) comprehensively; 
2) \textit{Comparison of Prompting Methods}, where we scrutinize the capabilities of GPT-4V alongside pretrained and instruction-tuned MLMs in both few-shot and zero-shot settings; 3) \textit{Analysis of Reasoning Ability}, where we probe the interpretable reasoning skills of GPT-4V and Gemini utilizing decision-making rationales.

Our findings could be summarized as follows:

\begin{itemize}[leftmargin=*,topsep=0.1em,itemsep=0.1em,parsep=0.1em]
    \item \textbf{MLMs exhibit varied reasoning abilities across knowledge domains}. Our analysis highlights a significant variance in the understanding and reasoning capabilities of MLMs across diverse knowledge categories, encompassing both commonsense and fine-grained world knowledge.

    \item \textbf{VQA with fine-grained world knowledge is challenging for GPT-4V and Gemini}. Our evaluation in Sec.~\ref{fine_grained} identifies four primary issues GPT-4V encounters in fine-grained world knowledge Q\&A: a) \textit{Reluctance to Respond Due to Insufficient Context}; b) \textit{Challenges in Recognizing Similar Objects (Visual Illusion)}; c) \textit{Inadequate Integration of Visual and Knowledge Dimensions}; d) \textit{Overreliance on Visual Clues, Overlooking Textual Queries}.
    
    \item \textbf{GPT-4V can handle composite image}.
    Our prompting approach involved feeding GPT-4V with a composite image containing contextual references for in-context learning. This technique, as evidenced by our experimental results (referenced in Table~\ref{tab:aokvqa_results}), enables GPT-4V to achieve heightened answer accuracy. In particular, the incorporation of contextual reference examples within the composite image enhances the quality of generated rationales, improving the decision-making interpretation of GPT-4V.
\end{itemize}
\begin{figure}[t]
    \centering
    \includegraphics[width=0.98\textwidth]{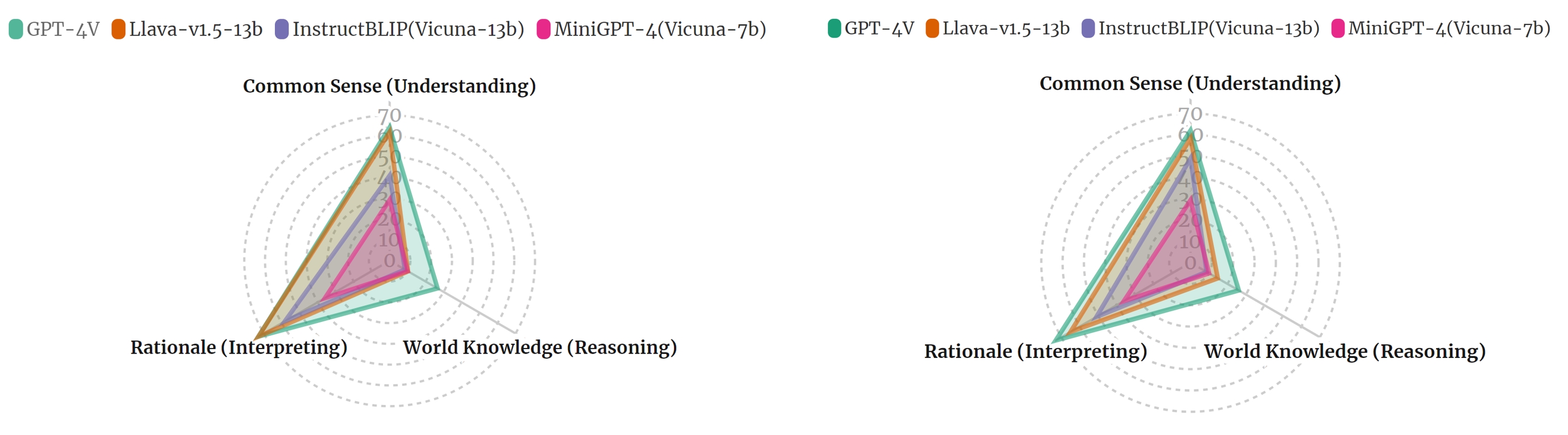}
    \caption{GPT-4V achieves SOTA performance in terms of common sense, fine-grained world knowledge, and decision-making rationale generation. The left and right parts represent the models with zero-shot and few-shot settings, respectively.}
    \label{fig:leida}
\end{figure}

\section{Related Work}

\textbf{Knowledge-based Visual Question Answering}. 
Visual Question Answering (VQA) is an interdisciplinary field that fuses vision and language processing to answer questions about images. A recent advancement in this area is knowledge-based VQA, which requires external information to answer open-domain visual questions. This is distinct from traditional VQA datasets that rely solely on information contained within the image. The earliest explicitly knowledge-based VQA datasets were KB-VQA~\cite{wang2015explicit} and FVQA~\cite{wang2017fvqa}. However, the knowledge needed to answer questions in these benchmarks was limited to specific, predefined information described as "closed" knowledge. In contrast, S3VQA~\cite{jainSelect} and OK-VQA~\cite{marino2019ok} datasets introduced questions that required "open"-domain knowledge, meaning that the information needed for answers was not drawn from a singular, specific source and could include commonly understood facts across various domains. To step forward in this direction, the recently proposed Wikipedia-based VQA dataset, INFOSEEK~\cite{chen2023can} focused on fine-grained entity knowledge for open-domain information-seeking questions.
Consequently, datasets like OK-VQA and INFOSEEK, which encompass extensive categories and scopes of knowledge, are particularly suitable for evaluating the capabilities of MLMs in open-domain VQA tasks. A-OKVQA~\cite{AOKVQA} is a new knowledge-based visual question answering benchmark, requiring a broad base of commonsense and world knowledge to answer. It consists of four knowledge types: Commonsense Knowledge about the world that humans learn from their everyday experiences; Visual Knowledge of concepts represented visually;
Knowledge bases Knowledge obtained from textbooks, Wikipedia, and other textual sources; Physical Knowledge about the physics of the world. Notably, A-OKVQA also features "rationale" annotations. These allow for a more granular examination of how knowledge-based VQA systems acquire and apply information to reasoning processes~\cite{li-etal-2023-neural}. Such annotations are pivotal for a deeper evaluation of MLMs' visual reasoning and knowledge integration capabilities, paving the way for future developments in the field.

\textbf{Multimodal Large Models}.
Recent advancements in the development of foundational models for vision and language have accelerated the creation of multimodal large models. In 2023, inspired by the release of multimodal large model GPT-4, several multimodal large models~\cite{GVT2023, ye2023mplug, lyu2023macaw, li2023lmeye, liu2023improved} were introduced and have demonstrated notable performance across a range of visual-language tasks. These models typically employ pretrained visual models to extract visual features and then integrate these features into the linguistic space of Large Language Models (LLMs) using a simple projection layer, such as a Linear Layer~\cite{merullo2022linearly, liu2023visual, li2023multi} or a Q-former~\cite{li2023blip2}. Subsequently, akin to the supervised fine-tuning approach used for LLMs, these systems are refined using diverse and high-quality multimodal instruction-following datasets~\cite{liu2023visual, zhu2023minigpt, ye2023mplug}, which include both human-labeled datasets for downstream tasks like Visual Question Answering (VQA) and VCR~\cite{VQA,zellers2019recognition} and datasets automatically generated by GPT-4. Although some multimodal benchmarks, such as MMBench~\cite{liu2023mmbench}, have been established to assess the capabilities of advanced MLMs and GPT-4V~\cite{li2023comprehensive}, a comprehensive evaluation of their performance on knowledge-intensive VQA scenario remains to be conducted.

\section{Experimental Setup}

\subsection{Dataset}

\begin{table}
\renewcommand\arraystretch{1.3}
\tabcolsep=0.11cm
\small
\caption{Dataset Statistics for Knowledge-based VQA. A: answer, R: rationale, DA: Direct Answer, KB: Knowledge Base, MC: Multiple Choice. 'NA' denotes the absence of rationales.}
\centering
\scalebox{0.96}{
\label{tab:dataset_table}
    \begin{tabular}{ccccccc}
        \toprule
         Dataset & Questions & Images & Rationale &Knowledge type & Ans type & Avg. length (Q/A/R)\\
        \midrule
        KB-VQA & 2,402 & 700 & $\times$ & fixed KB & DA & 6.8/2.0/NA \\
        FVQA & 5,826 & 2,190 & \checkmark & fixed KB & DA & 9.5/1.2/NA \\
        OK-VQA  & 14,055 & 14,031 & $\times$ & common & DA & 8.1/1.3/NA  \\
        S3VQA  & 7,515 & 7,515 & $\times$ & common & DA & 12.7/2.8/NA  \\
        VCR  & 290k & 99,904 & \checkmark & people actions & MC & 8.7/7.7/16.8\\
        INFOSEEK & 1.35M & 1.35M & $\times$ &  fine-grained/world & DA & 8.9/1.5/NA\\
        A-OKVQA & 24,903 & 23,692 & \checkmark & common/world & DA/MC & 8.8/1.3/11.0\\
        \hline
        Our Benchmark & 4,290 & 4,189 & \checkmark & common/fine-grained/ world & DA & 8.6/1.5/11.2\\
        \bottomrule 
    \end{tabular}}
\end{table}

\begin{figure}
    \centering
    \includegraphics[width=0.9\textwidth]{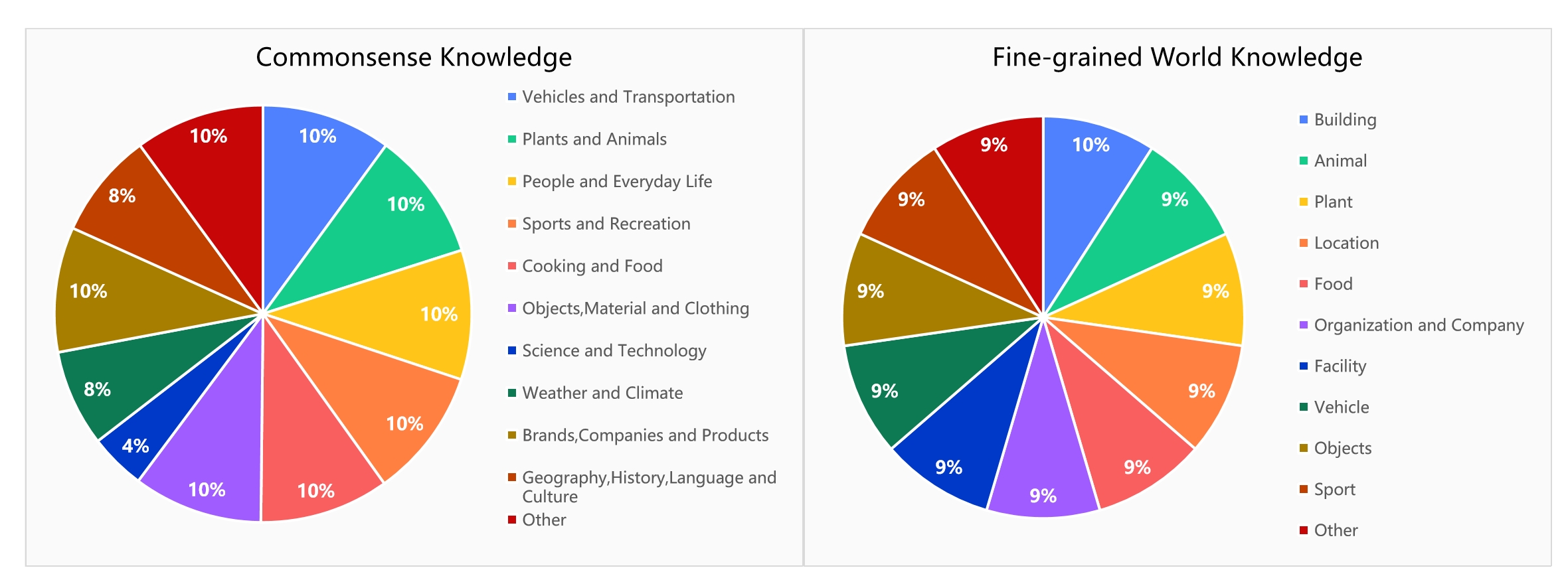}
    \caption{Breakdown of questions in terms of knowledge
categories for commonsense and world knowledge types. We show the percentage of questions falling into each of the 11 knowledge categories.}
    \label{fig:data_statics}
\end{figure}

Table~\ref{tab:dataset_table} shows the detailed statistics of knowledge-based VQA datasets. Our study acknowledges that different datasets incorporate varying forms of knowledge, and accordingly, employ diverse evaluation dimensions. To assess the abilities of MLMs (GPT-4V) in handling knowledge-intensive VQA tasks, we have meticulously selected three prominent and up-to-date knowledge-based VQA datasets: OK-VQA (commonsense knowledge), INFOSEEK (fine-grained World Knowledge), and A-OKVQA (decision-making rationales). To facilitate a nuanced comparison of these MLMs across a spectrum of knowledge domains and types, we construct our evaluation benchmark based on the testing or validation sets of the above datasets. First, based on manually annotated questions, we split various knowledge categories and randomly selected 150 questions per one. These chosen questions constitute our primary set of evaluation benchmarks. In scenarios where certain knowledge categories comprised fewer than 150 samples, all available samples were included to ensure representativeness. We present the detailed distribution and category-wise breakup of these samples in the last line of Table~\ref{tab:dataset_table} and detailed statistics of knowledge scope are shown in Figure~\ref{fig:data_statics}. Particularly for INFOSEEK containing fine-grained world knowledge, our methodology focused on extracting samples whose questions were distinct from those in the training set, including questions and entities unseen in the training set. This approach was deliberately chosen to mitigate the risk of the models being inadvertently trained on the validation data, thereby ensuring a more robust and unbiased evaluation of their performance in real-world, knowledge-driven scenarios.

Additionally, A-OKVQA stands out as it encompasses decision-related rationales, integrating multiple types of knowledge including Common Sense, World Knowledge, Visual Knowledge, and Physical Knowledge, all enhanced by manual annotation. This dataset serves a crucial role in our study: it enables us to rigorously assess the accuracy of decision-making processes. For this purpose, we utilize the whole validation set of A-OKVQA data set, providing a comprehensive decision-making rationale evaluation for MLMs.

In conclusion, our evaluation benchmark consists of various knowledge categories, covering more than ten knowledge categories, e.g., plants, animals, food, location, and others. Its knowledge types mainly focus on common sense and fine-grained world knowledge. In addition, we also introduce the decision-making rationales to assess how MLMs acquire and apply knowledge to reasoning processes. It significantly impacts the practical adoption of VQA systems in real-world scenarios. 

\subsection{Evaluation Method}

From Table~\ref{tab:dataset_table}, we observe that the mostly answers are short content with a few words and the rationale is sentence-level. Hence, we adopt the following three evaluation methods:

\textbf{Exact Matching}. The most straightforward method of evaluation is Exact Matching, which involves comparing the generative short answer of a VQA task directly against a set of pre-defined correct answers. This method assesses accuracy by checking if the generated answer exactly matches any of the answers in the reference set. 

\textbf{Automatic Rationale Evaluation}. This evaluation not only considers the answer's correctness but also how well the model's response explains the reasoning or logic behind the answer. We employ generative metrics such as BLEU, CIDER, and METEOR to evaluate the linguistic quality and relevance of these rationale statements in comparison to a reference set. These metrics, while originally designed for generative tasks like machine translation, can provide insights into the coherence and fluency of the generated explanations, augmenting the evaluation of answer correctness.

\textbf{Human Evaluation}. In addition to automatic evaluation, human judgment is unparalleled in understanding context, nuance, and the subtleties of natural language, which are often essential for evaluating the coherence and relevance of the explanations provided by VQA systems. The interpretability of rational sentences is best gauged against human cognition to ensure that the explanations align with human reasoning patterns.
Human evaluators will assess the generated rationale sentences from the following three  aspects:
\begin{itemize}
    \item Consistency: Evaluators should determine if the rationale logically follows from the visual content and the question posed, maintaining a consistent narrative throughout.
    \item Sufficiency: The evaluation must ascertain if the rationale addresses the question adequately, providing enough information without superfluous details.
    \item Factual Correctness: The rationale must be not only plausible but also factually correct, aligning with the actual content of the image and the knowledge base MLMs draw upon.
\end{itemize}

\section{Assessing MLMs in Commonsense Knowledge Q\&A}
Commonsense knowledge, understood as fundamental and evident truths, poses a significant challenge for artificial intelligence (AI) as computers inherently lack the innate understanding that humans possess. AI systems often falter when faced with tasks requiring commonsense reasoning, such as understanding the necessity of wearing a coat in cold weather. A key objective within AI research is to devise ways to imbue computers with commonsense knowledge, which is crucial for enabling smooth and natural interactions between AI systems and humans. In the following sections, we examined the GPT-4V’s capability for open-ended VQA with commonsense knowledge using distinct prompting strategies.

\subsection{Evaluation}

\subsubsection{Dataset} 
As the left part is shown in Figure~\ref{fig:data_statics}, we use a subset of the OK-VQA testing set, which consists of 11 knowledge categories such as Plants and Animals, Cooking and Food, Science and Technology, etc. The total number of samples is 1495. Each of the major categories: Vehicles and Transportation (VT), Plants and Animals (PA), Other, People and Everyday Life (PEL), Sports and Recreation (SR), Cooking and Food (CF), Objects, Material and Clothing (OMC) are well-represented with 150 entries each, ensuring a balanced variety across these common domains. However, certain categories like Science and Technology (ST), Weather and Climate (WC), and Geography, History, Language, and Culture (GHLC) are less represented with 65, 112, and 123 entries respectively, which suggests potential areas for expansion. Brands, Companies, and Products (BCP) also have a substantial share with 145 entries, indicating a good spread across both commercial and natural aspects. Overall, the dataset is enough comprehensive to evaluate MLMs on commonsense knowledge visual Q\&A.

\begin{figure}[t]
    \centering
    \includegraphics[width=0.98\textwidth]{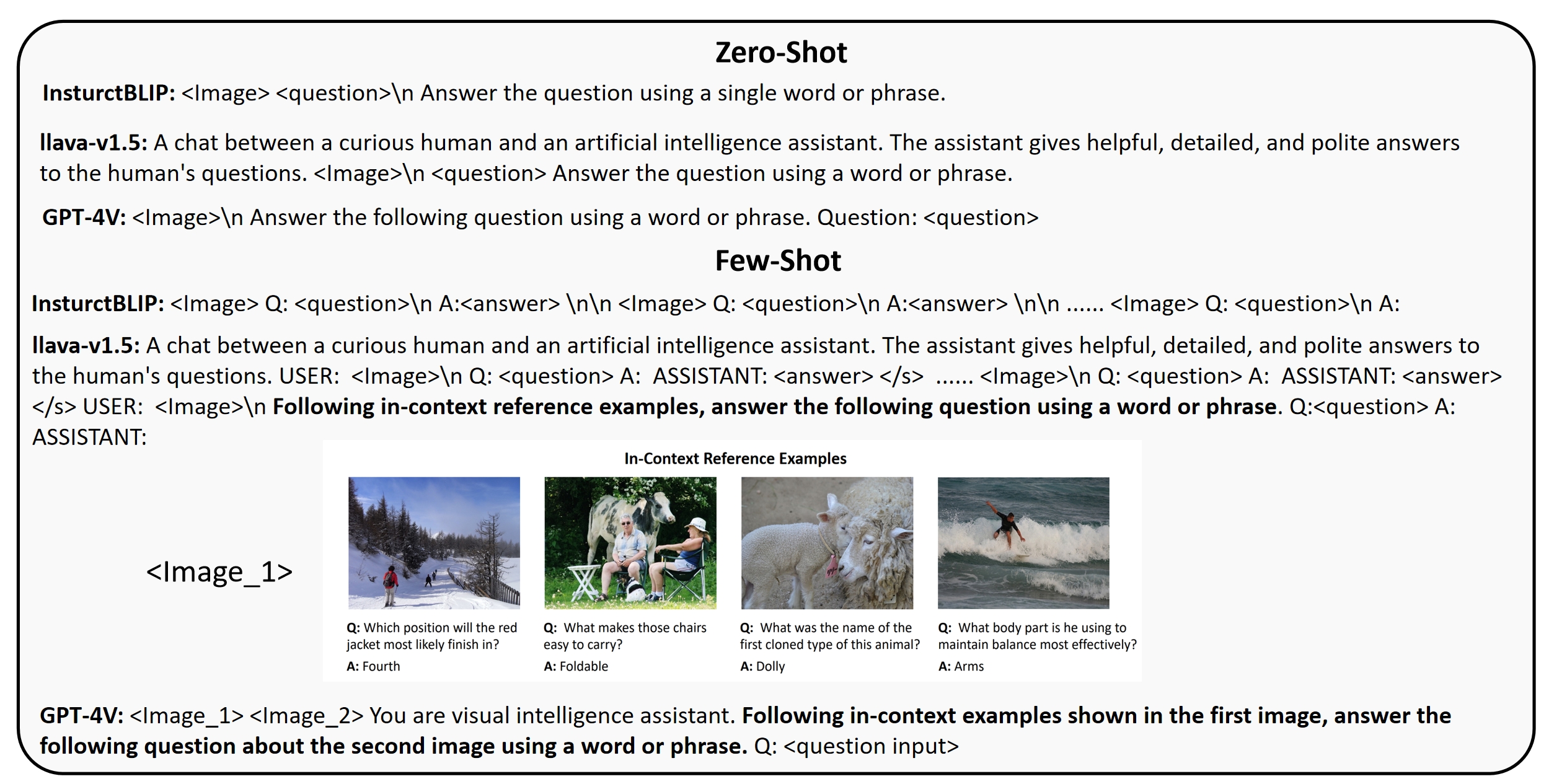}
    \caption{Prompting Technical for MLMs. Due to the image input limitation of GPT-4V, we employ its in-context learning capability by inputting a composite image that contains four reference samples.}
    \label{fig:ok-vqa-prompt}
\end{figure}

\begin{table}[t]
\renewcommand\arraystretch{1.3}
\tabcolsep=0.14cm
\centering
\scriptsize
\caption{Benchmark results on VQA with Commonsense Knowledge. ${\clubsuit}$ refers to that the corresponding model is trained using the training set of OK-VQA. Vehicles and Transportation (VT); Brands, Companies and Products (BCP); Objects, Material and Clothing (OMC); Sports and Recreation (SR); Cooking and Food (CF); Geography, History, Language and Culture (GHLC); People and Everyday Life (PEL); Plants and Animals (PA); Science and Technology (ST); Weather and Climate (WC); and Other. }
\label{tab:okvqa_results}
\begin{tabular}{l|c|cccccccccccc}
\toprule
Model & Avg. & VT & BCP & OMC & SR & CF & GHLC & PEL & PA & ST & WC & Other \\
\hline
MLP$^{\clubsuit}$ & 20.67 & 21.33 & 15.81 & 17.76 & 24.69 & 21.81 & 11.91 & 17.15 & 21.33 & 19.29 & 29.92 & 19.81 \\
ArticleNet (AN)$^{\clubsuit}$ & 5.28 & 4.48 & 0.93 & 5.09 & 5.11 & 5.69 & 6.24 & 3.13 & 6.95 & 5.00 & 9.92 & 5.33 \\
BAN$^{\clubsuit}$ & 25.17 & 23.79 & 17.67 & 22.43 & 30.58 & 27.90 & 25.96 & 20.33 & 25.60 & 20.95 & 40.16 & 22.46 \\
MUTAN$^{\clubsuit}$ & 26.41 &25.36 &18.95 &24.02 &33.23& 27.73 &17.59 &20.09 &30.44 &20.48 &39.38 &22.46 
\\
BAN + AN$^{\clubsuit}$ &25.61 &24.45 &19.88 &21.59 &30.79 &29.12 &20.57 &21.54 &26.42 &27.14 &38.29 &22.16
\\
MUTAN + AN$^{\clubsuit}$ &27.84 &25.56 &23.95 &26.87& 33.44 &29.94 &20.71 &25.05 &29.70 &24.76 &39.84 &23.62\\
\hline
MiniGPT-4 (Vicuna-7b) & 29.31 & 28.67 & 31.03 & 26.0 & 28.0 & 25.33 & 38.21 & 22.67 & 29.33 & 29.23 & 31.25 & 34.0\\
MiniGPT-4 (V-4-shot) & 25.96 & 22.67 & 26.21 & 26.0 & 22.0 & 27.33 & 29.27 & 24.67 & 21.33 & 21.54 & 35.72 & 29.33\\
BLIP-2 (FlanT5-xxl) & 39.06 & 30.67 & 34.48 & 38.0 & 40.67 & 34.0 & 42.28 & 39.33 & 41.33 & 44.62 & 50.0 & 40.67\\
BLIP-2 (Flan-4-shot) & 25.95 & 22.0 & 26.90 & 25.33 & 22.67 & 20.67 & 31.71 & 20.02 & 22.0 & 24.62 & 36.10 & 36.0\\
InstructBLIP$^{\clubsuit}$ (Vicuna-13b) & 41.02 & 34.00 & 52.41 & 37.33 & 51.33 & 33.33 & 46.34 & 31.33 & 38.67 & 32.30 & 49.11 & 43.33\\
InstructBLIP$^{\clubsuit}$ (V-4-shot) & 48.90 & 41.33 & 60.0 & 47.33 & 49.33 & 48.67 & 52.85 & 38.67 & 47.33 & 44.62 & 51.79 & 55.33 \\
InstructBLIP$^{\clubsuit}$ (FlanT5-xxl) & 47.96 & 44.66 & 51.03 & 48.67 & 48.0 & 43.33 & 51.22 & 47.33 & 42.0 & 55.38 & 58.04 & 45.33 \\
InstructBLIP$^{\clubsuit}$ (Flan-4-shot) & 41.67 & 34.0 & 53.80 & 43.33 & 43.33 & 38.67 & 50.41 & 37.33 & 35.33 & 44.62 & 37.5 & 42.67 \\
llava-v1.5-13b$^{\clubsuit}$ (0-shot) & 61.93 & 58.67 & 61.38 & 57.33 & 63.33 & 62.0 & 68.29 & 60.0 & 58.67 & 64.62 & 70.53 & 61.33\\
llava-v1.5-13b$^{\clubsuit}$ (4-shot) & 58.32 & 54.0 & 63.75 & 54.0 & 61.33 & 54.0 & 62.60 & 54.0 & 57.33 & 64.62 & 62.5 & 60.0 \\
\hline
Gemini (0-shot) & 62.27 & 55.33 & 63.45 & 72.0 & 65.33 & 61.33 & 58.53 & 60.0 & 62.67 & 64.62 & 60.71 & 64.67 \\
GPT-4V (0-shot) & \textbf{64.28} & 55.33 & 61.37 & 68.0 & 66.66 & 68.67 & 61.78 & 61.33 & 64.67 & 70.77 & 63.40 & 68.0 \\
GPT-4V (4-shot) & 62.01 & 57.33 & 60.0 & 69.33 & 64.67 & 62.67 & 60.97 & 57.33 & 58.67 & 67.69 & 61.61 & 64.67\\
\bottomrule
\end{tabular}
\end{table}

\subsubsection{Results and Analysis}

\textbf{Comparative models}. The corresponding evaluation results are shown in Table~\ref{tab:okvqa_results}. The comparative models consist of the traditional small neural network models (such as MLP, Article Net, and others) and multimodal large models (BLIP-2~\cite{li2023blip2}, MiniGPT-4~\cite{zhu2023minigpt}, InstructBLIP~\cite{liu2023visual}, and Llava-v1.5~\cite{liu2023improved}, GPT-4V~\cite{gpt4}). Small neural networks have been trained on the training set of OK-VQA. ${\clubsuit}$ refers to that the corresponding model is trained using the training set of OK-VQA.

\textbf{Prompt Method}. 
After extensive research into various methods of providing instructions to open-source MLMs and GPT-4V, we have identified an optimal prompt type that enhances the model's comprehension and facilitates answer accuracy. The selection of this particular prompt was made after a thorough evaluation of its effectiveness. Since the response is usually a word or phrase and different MLMs employ various prompting methods during training, we take the prompting strategies for different models, which are presented in Figure~\ref{fig:ok-vqa-prompt}. \textit{Since GPT-4 can only process up to four images at a time, we suggest using a visual input approach to prompt the model to produce relevant responses based on the provided in-context reference examples in a composite image.} The four in-context images are divided into commonsense, physical knowledge, world knowledge, and visual knowledge, which are shown in Figure~\ref{fig:case_intro}.

\textbf{Comparison and Analysis}. 
Table~\ref{tab:okvqa_results} indicates that Llava-v1.5-13b is the most proficient open-source MLM in common sense visual question answering, yet it is much lower than GPT-4V, especially in 'Science and Technology', 'Objects, Material and Clothing', and 'Other' domains. In addition, fine-tuning MLMs using relevant common sense VQA data will greatly improve the performance of MLMs on common sense visual Q\&A. The application of 4-shot in-context learning affects the common sense reasoning performance of some MLMs, with open-source models being more sensitive to this impact than GPT-4V. It may be attributed to the training technical or few-shot examples we used, which affects the in-context learning performance of MLMs. The 4-shot prompting method we used is effective for improving the performance of GPT-4V, e.g., it could improve the performance on VT and OMC knowledge categories.

Analyzing the performance across various knowledge categories reveals a long-tail effect in the application of internal knowledge by MLMs. For instance, GPT-4V's performance in 'Science and Technology' (ST) significantly exceeds its performance in 'Vehicles and Transportation' (VT), scoring 70.77 compared to 55.33. Overall, while open-source MLMs show promise, GPT-4V remains significantly ahead in a wide range of common sense categories.

\begin{figure}
    \centering
    \includegraphics[width=1.0\textwidth]{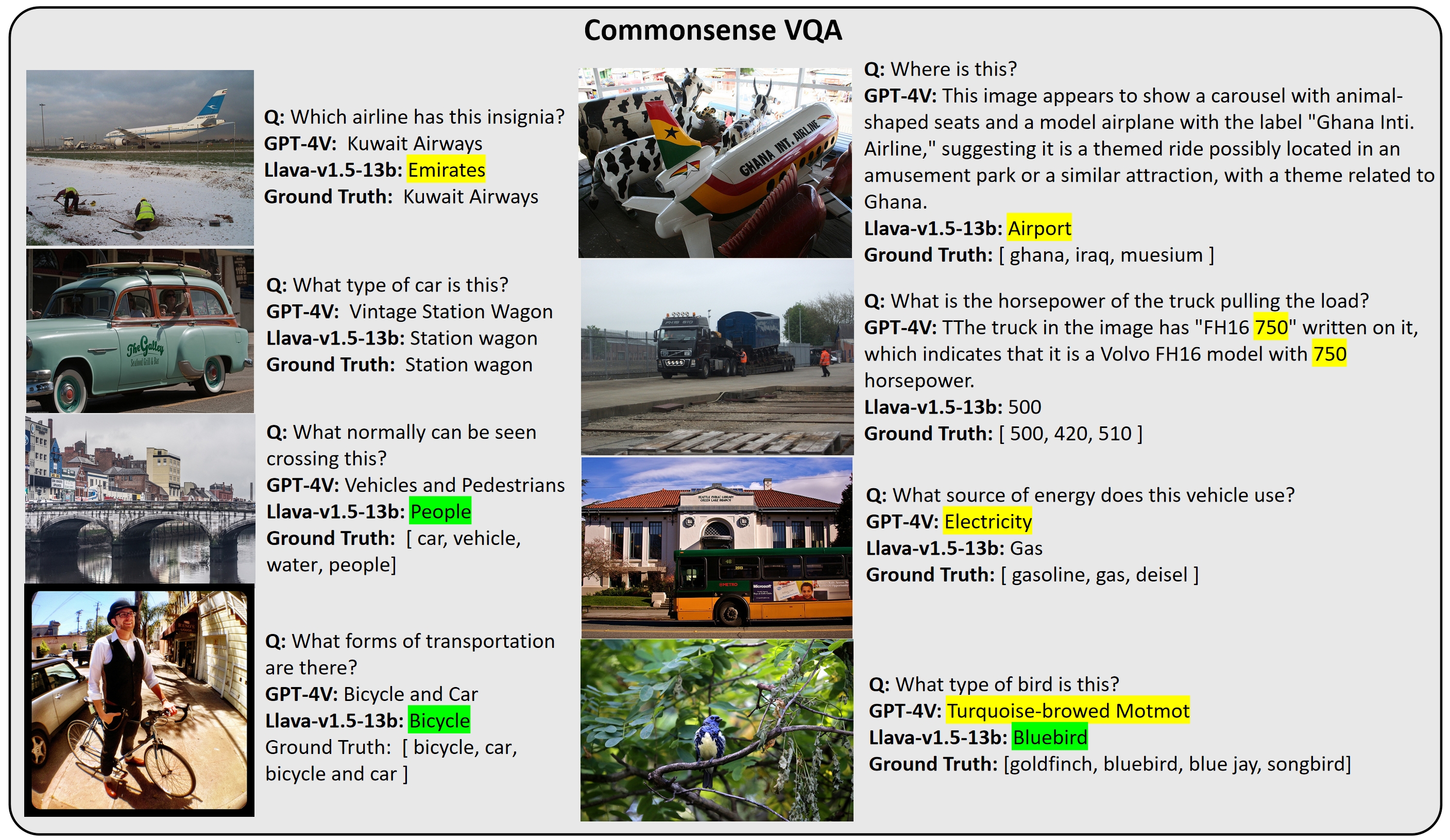}
    \caption{The illustration of cases generated by GPT-4V and Llava-v1.5-13b on commonsense knowledge. Words with \textcolor[RGB]{255,216,0}{yellow} and \textcolor{green}{green} background refer to the incorrect answer and inadequate responses, respectively. \textit{GPT-4V tends to generate more detailed and accurate answers yet has the visual illusion problem}.}
    \label{fig:commensense_vqa}
\end{figure}

\textbf{Case Study}. Figure~\ref{fig:commensense_vqa} presents a side-by-side comparison of responses generated by GPT-4V and Llava-v1.5-13 (trained with the training set of OK-VQA). GPT-4V tends to generate more comprehensive and precise responses than Llava-v1.5-13b. However, it is not without errors; GPT-4V occasionally provides incorrect answers due to misinterpretations of visual content. This is particularly evident in the two examples displayed in the middle of the right column, where GPT-4V misidentifies key elements in the images, such as mistaking recognition of an FH16 510 and misunderstanding the context of wires near buses. When it comes to recognizing animals and plants, GPT-4V demonstrates high accuracy according to the results shown in Table~\ref{tab:okvqa_results}. Despite this, it struggles to distinguish between certain similar species, as illustrated by the blue songbird example (the last case in the right column). After our continuous testing, we found that GPT-4V is sometimes able to generate detailed inference processes as the top example shown in the right column, especially when it gives an uncertain response.

\subsection{Conclusion}

In this section, we evaluate the performance of GPT-4V and various advanced open-source multi-modal large models on common-sense VQA tasks. The experimental results and the examples showcased indicate that GPT-4V outperforms its peers, delivering more detailed and comprehensive answers. \textit{We note that multi-modal large models, when fine-tuned with common-sense questions and answers, exhibit marked improvements in performance in this domain}. However, when operating under a few-shot input setting, there is no substantial enhancement in the performance of most multi-modal large models, including GPT-4V. While challenges in example selection may contribute, GPT-4V's performance gains in certain knowledge categories suggest the feasibility of refining the model through in-context reference examples in a composite image.

\section{Evaluating MLMs on Fine-grained World Knowledge}

Compared to commonsense knowledge, world knowledge refers to specific information about the world which could include facts, information, and data about geography, history, science, current events, and more. This kind of knowledge is more detailed and specific. For instance, knowing the capital of France is Paris, understanding the implications of a new scientific discovery, or being aware of the current political climate. World knowledge allows the AI to answer factual and specific queries, is critical for MLMs to answer the information-seeking questions. In the following sections, we mainly investigate the capability of current MLMs to answer visual questions relevant to fine-grained world knowledge. This type of VQA needs MLMs to recognize the visual content and connect visual clues with the stored knowledge.

\subsection{Dataset}
In this section, we evaluate the performance of GPT-4V, focusing on its ability to handle fine-grained knowledge across various categories. As illustrated in the right section of Figure~\ref{fig:data_statics}, our analysis utilizes a subset of the INFOSEEK validation dataset. INFOSEEK, detailed in Chen et al. (2023)~\cite{chen2023can}, is a dataset grounded in Wikipedia knowledge and comprises approximately 1.35 million question-image pairs suitable for visual question-answering (VQA) tasks.
For a nuanced assessment, we categorized the validation set of INFOSEEK into 11 distinct knowledge domains based on the types of questions posed. From each of these categories, we randomly selected 150 samples, amounting to a total of 1650 samples for our evaluation set. This approach ensures a balanced representation of each category and facilitates a fair comparison of the GPT-4V's performance across fine-grained knowledge domains.

\begin{figure}
    \centering
    \includegraphics[width=0.98\textwidth]{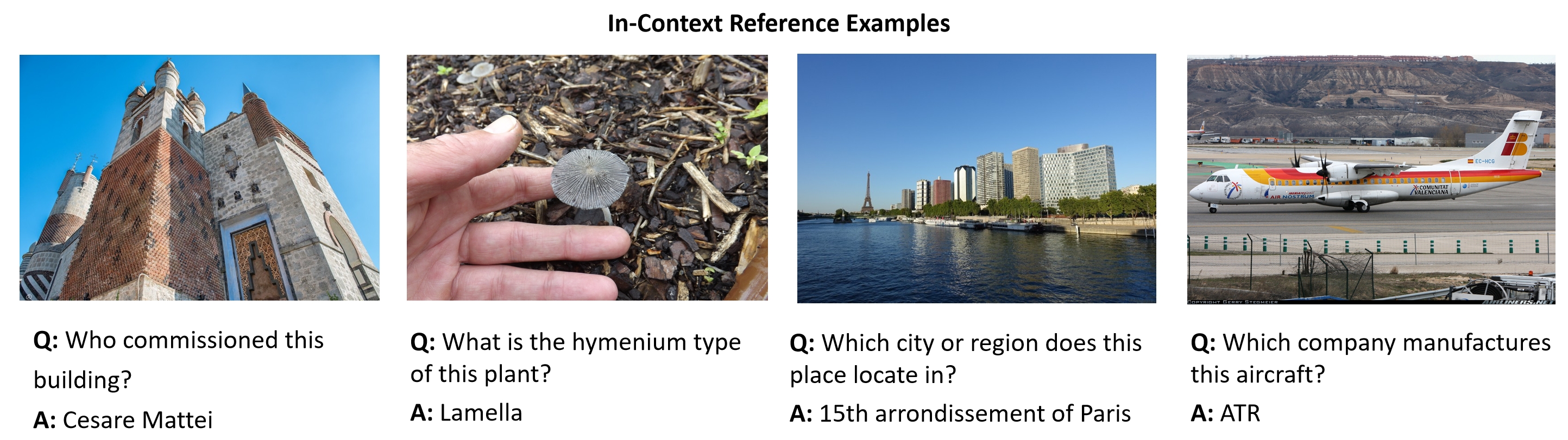}
    \caption{The illustration of in-context reference examples for MLM evaluation on fine-grained world knowledge. Other prompting methods are identical to commonsense VQA.}
    \label{fig:info-vqa-prompt}
\end{figure}

\subsection{Results and Analysis}
\label{fine_grained}
\textbf{Comparing Models}. We mainly evaluate MLMs (BLIP-2~\cite{li2023blip2}, MiniGPT-4~\cite{zhu2023minigpt}, InstructBLIP~\cite{liu2023visual}, and Llava-v1.5~\cite{liu2023improved}, GPT-4V~\cite{gpt4}) on fine-grained world knowledge with 0-shot and 4-shot input setting.

\textbf{Prompt Method}. We use the same prompting method for MLMs, as illustrated in Figure~\ref{fig:ok-vqa-prompt}. Due to the differences in the data types of fine-grained world knowledge and commonsense, we selected four image and question-answer pairs from the INFOSEEK data set (not included in our evaluation set) as new in-context reference examples, which are presented in Figure~\ref{fig:info-vqa-prompt}. 

\begin{table}[t]
\renewcommand\arraystretch{1.3}
\tabcolsep=0.12cm
\centering
\scriptsize
\caption{Benchmark results on VQA with fine-grained world knowledge. OC refers to Organization and Company. ${\clubsuit}$ refers to that the corresponding model is trained using the training set of OK-VQA.}
\label{tab:infoseek_results}
\begin{tabular}{l|c|cccccccccccc}
\toprule
Model & Avg. & Building & Animal & Plant & Location & Food & OC & Facility & vehicle & Objects & Sport & Other \\
\hline
MiniGPT-4 (Vicuna-7b) & 10.03 & 7.33 & 6.66 & 5.33 & 10.0 & 24.67 & 4.0 & 7.33 & 18.67 & 6.67 & 14.0 & 8.67\\
MiniGPT-4 (V-4-shot) & 5.58 & 2.67 & 0.67 & 4.67 & 5.33 & 10.0 & 4.0 & 6.0 & 11.33 & 4.0 & 6.0 & 6.67\\
BLIP-2 (FlanT5-xxl) & 10.67 & 8.7 & 2.67 & 4.0 & 16.0 & 14.0 & 9.33 & 16.0 & 28.0 & 2.0 & 9.33 & 7.33 \\
BLIP-2 (Flan-4-shot) & 9.44 & 5.3 & 2.0 & 13.33 & 14.0 & 8.0 & 12.0 & 10.67 & 28.67 & 6.67 & 13.33 & 6.67\\
InstructBLIP$^{\clubsuit}$ (Vicuna-13b) & 8.50 & 3.3 & 2.0 & 1.33 & 10.0 & 10.67 & 6.0 & 4.67 & 26.67 & 2.67 & 20.67 & 5.33\\
InstructBLIP$^{\clubsuit}$ (V-4-shot) & 8.75 & 5.3 & 2.0 & 2.0 & 8.0 & 15.33 & 4.67 & 5.33 & 16.67 & 5.33 & 24.67 & 6.67\\
InstructBLIP$^{\clubsuit}$ (FlanT5-xxl) & 8.37 & 4.0 & 5.33 & 2.0 & 8.67 & 8.0 & 8.0 & 8.0 & 28.0 & 5.34 & 8.67 & 6.0 \\
InstructBLIP$^{\clubsuit}$ (Flan-4-shot) & 5.65 &  1.3 & 2.0 & 1.33 & 2.67 & 18.67 & 7.33 & 2.67 & 15.33 & 4.67 & 2.67 & 3.33\\
llava-v1.5-13b$^{\clubsuit}$ (0-shot) & 10.22 & 11.33 & 16.67 & 0.0 & 24.67 & 6.0 & 0.7 & 10.67 & 26.0 & 5.3 & 0.13 & 10.0\\
llava-v1.5-13b$^{\clubsuit}$ (4-shot) & 14.73 & 10.67 & 4.0 & 6.67 & 16.0 & 22.67 & 16.0 & 4.67 & 32.67 & 7.33 & 23.33 & 18.0 \\
\hline
Gemini (0-shot) & 26.48 & 27.33 & 11.33 & 6.66 & 29.33 & 36.0 & 8.66 &  30.66 & 33.33 & 12.67 & 72.0 & 23.33 \\ 
GPT-4V (0-shot) & \textbf{26.62} & 18.67 & 10.28 & 12.60 & 17.44 & 46.67 & 19.33 & 29.33 & 33.33 & 17.69 & 57.33 & 22.53\\
GPT-4V (4-shot) & 26.10 & 20.0 & 7.47 & 8.89 & 23.49 & 40.0 &20.67 & 30.0 & 32.67 & 13.08 & 58.67 & 23.24\\
\bottomrule
\end{tabular}
\end{table}

\begin{figure}
    \centering
    \includegraphics[width=0.98\textwidth]{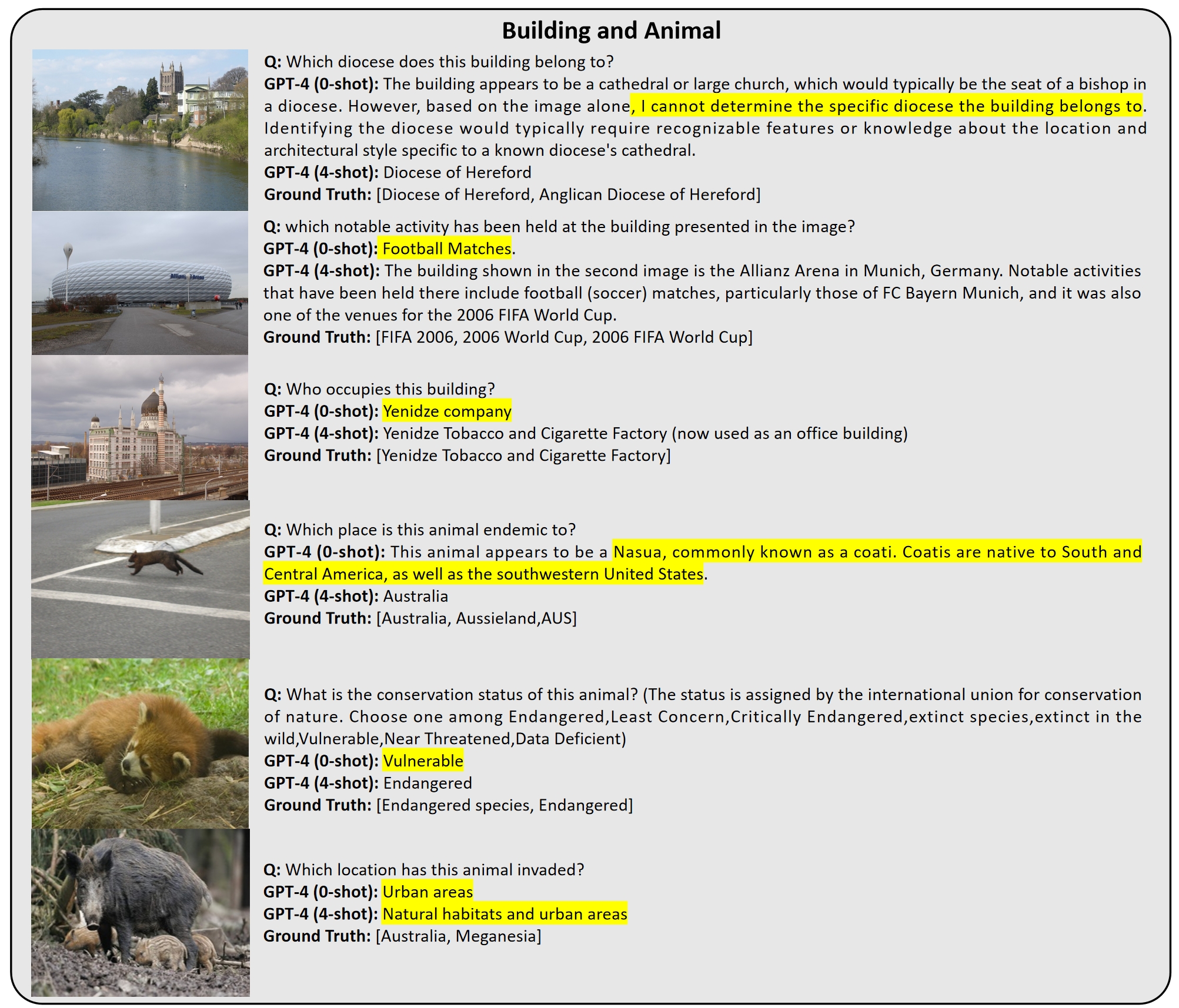}
    \caption{The upper part evaluates a building's architectural style and purpose, with clarifications on its actual diocese and notable events hosted. The lower section assesses the habitat, conservation status, and invasive presence of animals. GPT-4V struggles with recognizing similar objects (vision illusion)}
    \label{fig:infoseek_1}
\end{figure}
\begin{figure}
    \centering
    \includegraphics[width=0.98\textwidth]{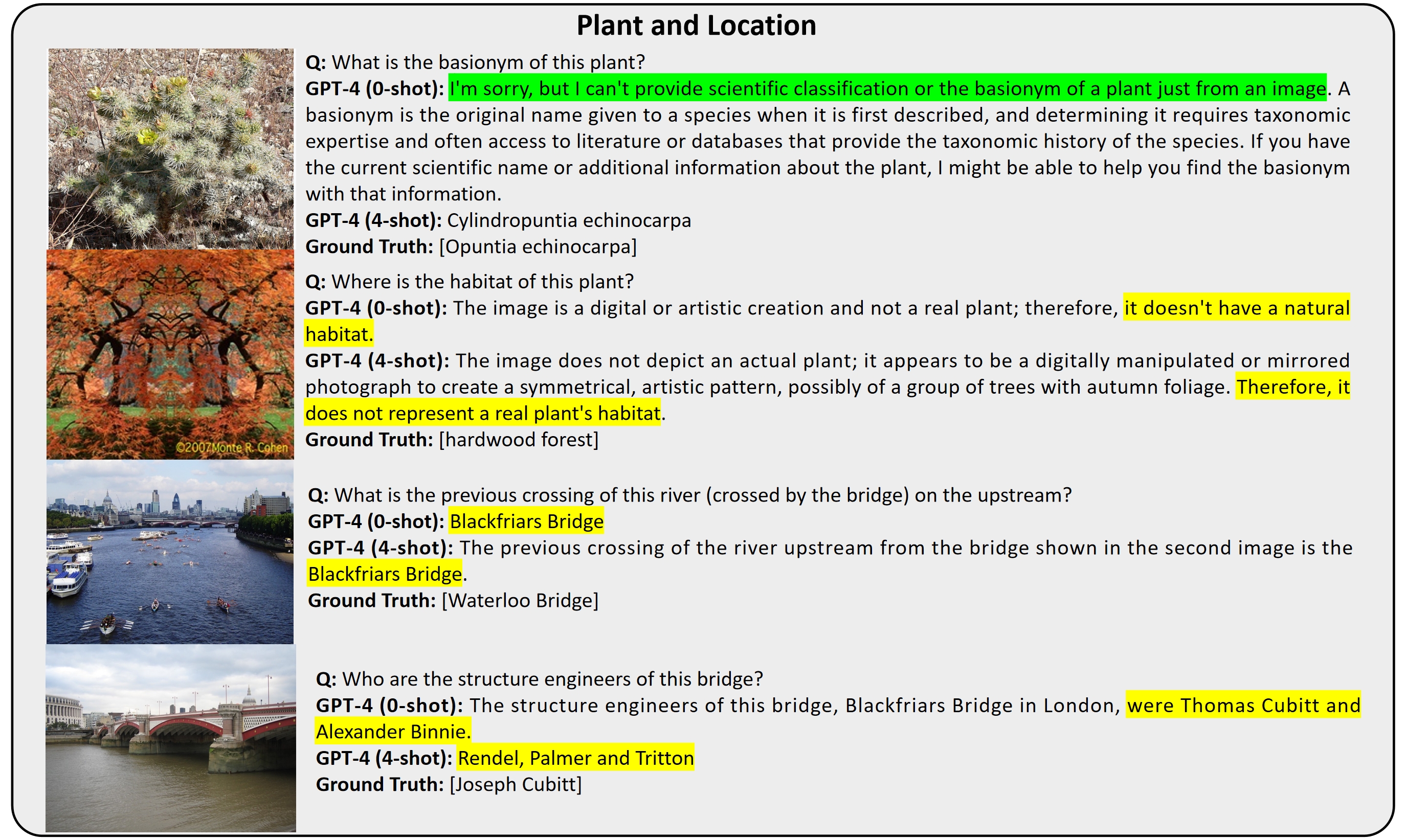}
    \caption{It showcases a misidentified plant species and its non-existent habitat, along with the historical details regarding a bridge's previous crossing and its engineering attribution. GPT-4V refuses to answer the question when the image can not provide sufficient information, over-relying the visual clues while answering the knowledge-intensive visual questions.}
    \label{fig:infoseek_2}
\end{figure}
\begin{figure}
    \centering
    \includegraphics[width=0.98\textwidth]{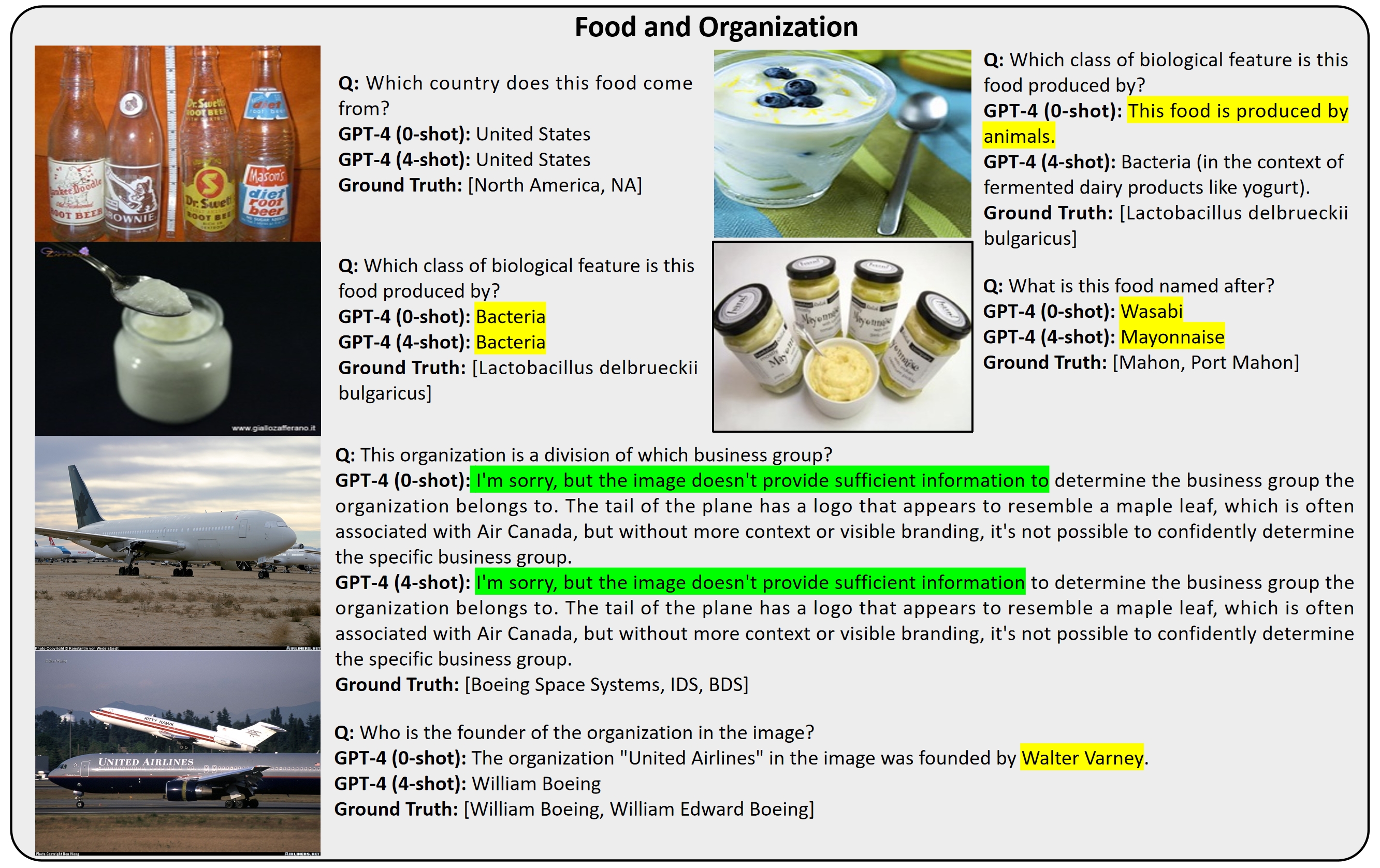}
    \caption{This figure contrasts GPT-4V generated responses with factual data regarding the origin and biological class of certain foods, as well as the correct historical naming and corporate lineage of an airline.}
    \label{fig:infoseek_3}
\end{figure}
\begin{figure}
    \centering
    \includegraphics[width=0.98\textwidth]{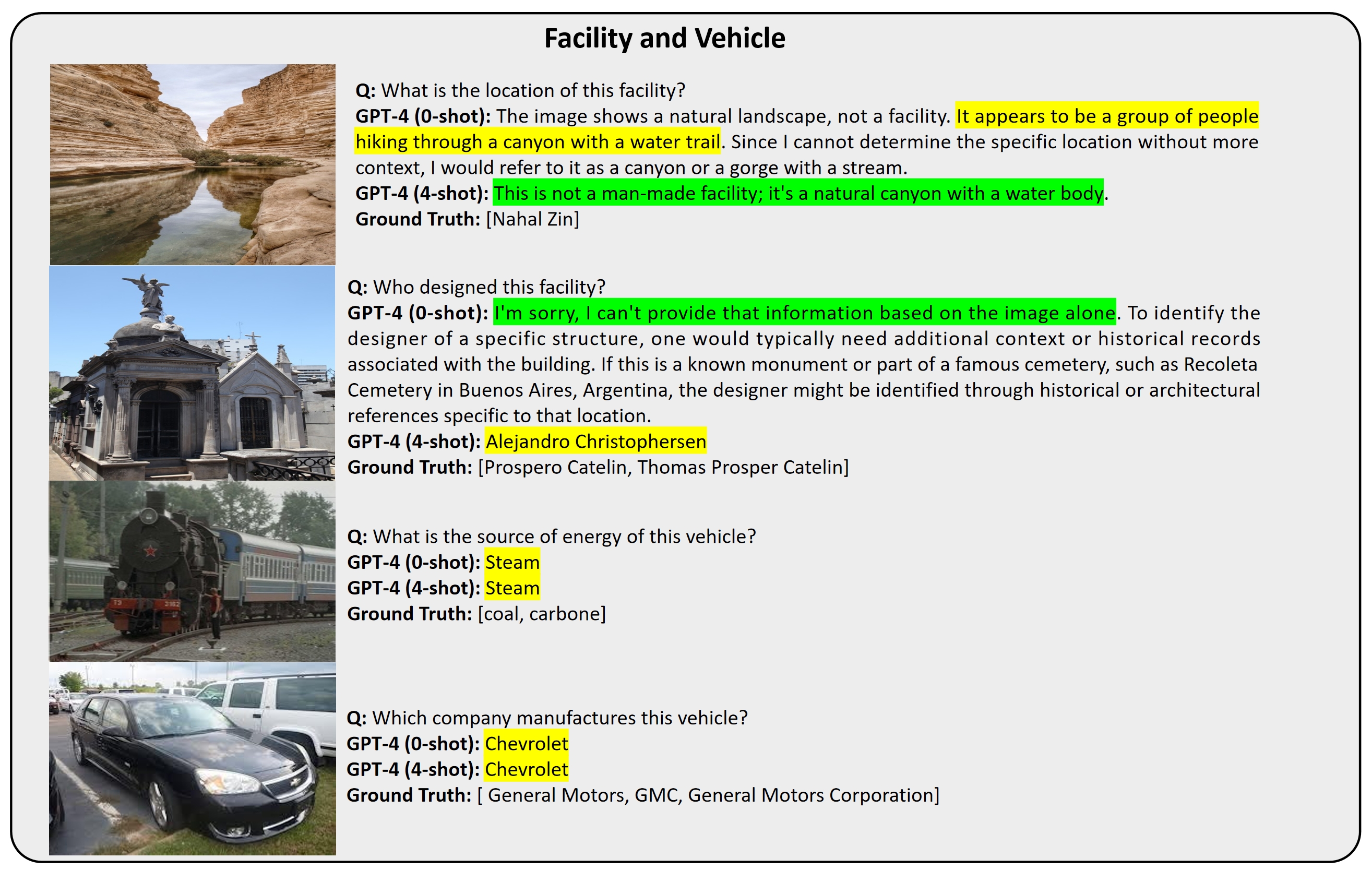}
    \caption{It presents a comparative analysis of responses from GPT-4 when asked to identify and provide details about various objects and locations depicted in images. The exercise illustrates the model's capabilities and limitations in interpreting visual information without additional context.}
    \label{fig:infoseek_4}
\end{figure}
\begin{figure}
    \centering
    \includegraphics[width=0.98\textwidth]{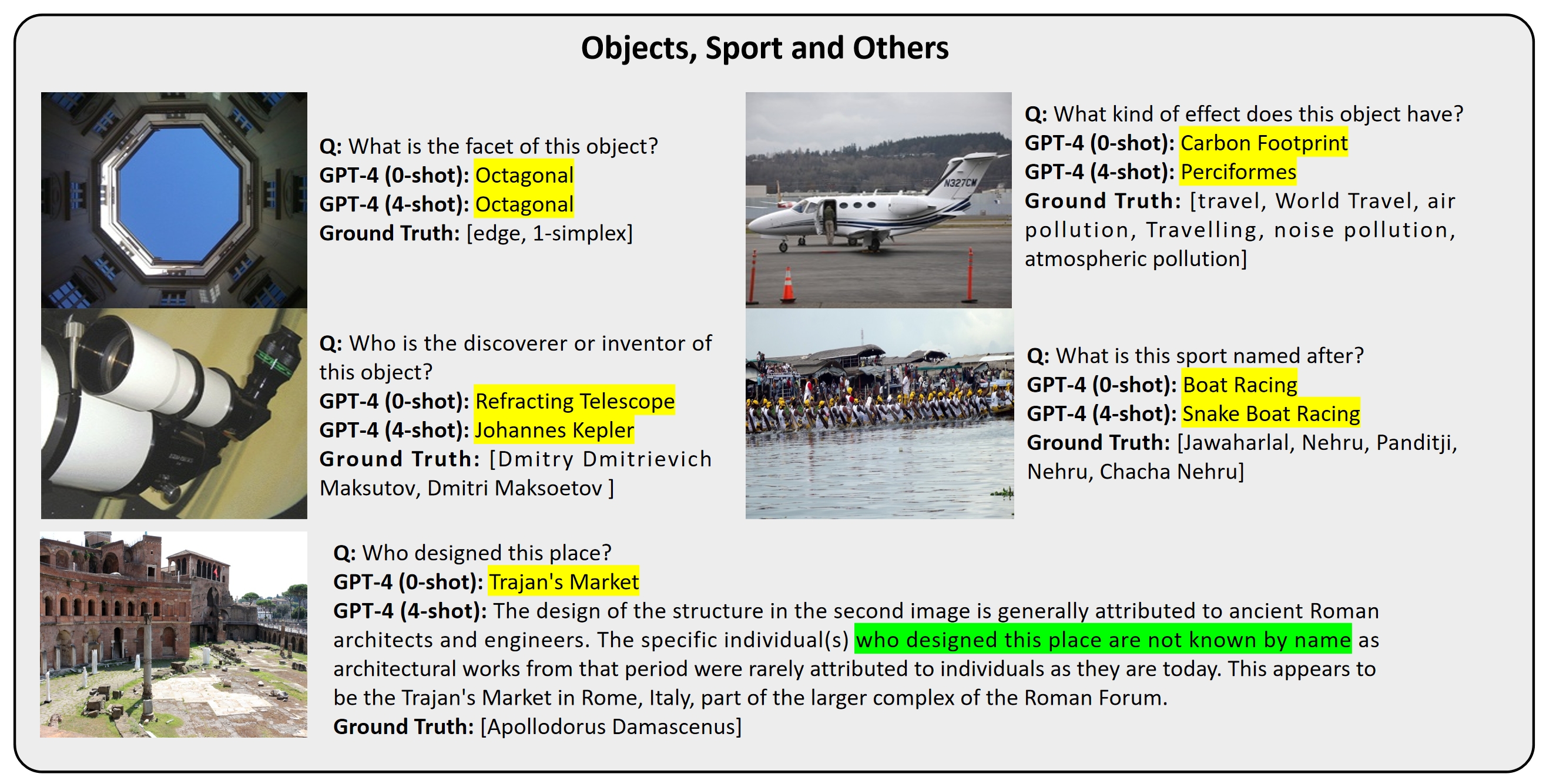}
    \caption{It includes assessments of a geometric structure, the inventor of a telescope, the environmental impact of an aircraft, and the historical context of an ancient Roman market, as well as the origins of a traditional boat race. The queries highlight the model's attempt to infer and deduce information from images and accompanying questions, demonstrating the challenges and intricacies of GPT-4V understanding and historical attribution.}
    \label{fig:infoseek_5}
\end{figure}

\textbf{Comparison and Analysis}. The experimental results are presented in Table~\ref{tab:infoseek_results}. 
Analysis reveals that GPT-4V, despite being a highly advanced multi-modal large model, attains an average accuracy of under 30\%. In comparison, open-source MLMs show a lower accuracy, not exceeding 15\%. This suggests a notable superiority of GPT-4V over open-source MLMs in handling fine-grained world knowledge. It is important to observe that all evaluated MLMs, regardless of their sophistication, demonstrate limited effectiveness in accurately addressing visual questions that require detailed world knowledge.
Notably, current open-source MLMs, which integrate pre-trained Language Large Models (LLMs) and visual encoders, inherit extensive knowledge from sources such as Wikipedia during their pre-training phase.  However, \textit{the performance of these MLMs on this single-hop visual-knowledge reasoning task suggests that they have not effectively bridged the gap between vision and fine-grained entity knowledge.} 

Additionally, we have observed a significant performance gap among MLMs across different knowledge categories, resulting in a severe long-tail phenomenon. For instance, the accuracy of Llava-v1.5-13b (zero-shot) on Plant is zero, yet reaches 26.0\% accuracy on Vehicle. Similarly, GPT-4V achieves an accuracy of 57.33\% on Sport, but only a 10.28\% accuracy on Animal, highlighting a substantial discrepancy of 47\%. An analysis of all models across various knowledge domains indicates that while they perform reasonably well in areas like Sport, Vehicle, and Animal, their effectiveness in the other eight knowledge domains is markedly lower. \textit{Therefore, the ability of current MLMs (including GPT-4V) to answer complex information-seeking questions requires further improvement.}

\textit{Moreover, a comparison between the zero-shot and few-shot input settings of the model substantiates the success of the visual in-context learning approach implemented in GPT-4V}. For instance, there has been an improvement in knowledge categories such as Building, Location, Sport, OC, and more. However, a decline in accuracy in certain categories may be linked to the specific in-context examples used. As illustrated in Figure~\ref{fig:ok-vqa-prompt}, our way of using the composite image, compared to traditional in-context sample inputs, helps reduce the length of context input tokens, thereby boosting the efficiency of the N-shot prompting approach. This efficiency gain is contingent on the multi-modal large models, like GPT-4V, inherent robust visual comprehension capabilities.

\textbf{Case Study}.
Figures \ref{fig:infoseek_1}-\ref{fig:infoseek_5} illustrate various error instances of GPT-4V across 11 knowledge categories in fine-grained world knowledge Q\&A. These errors are color-coded for clarity: a \textcolor[RGB]{255,216,0}{yellow} background denotes an incorrect response, while a \textcolor{green}{green} background signifies a refusal to answer. Our analysis identifies four primary issues GPT-4V encounters in VQA tasks involving detailed world knowledge:
\begin{itemize}
    \item \textit{Reluctance to Respond Due to Insufficient Context}. The green words shown in Figures~\ref{fig:infoseek_2},~\ref{fig:infoseek_3}, and ~\ref{fig:infoseek_4} indicate that GPT-4V often refuses to answer information-seeking questions when the image cannot provide sufficient information related to the question. GPT-4V tends to be overly cautious, often opting not to respond to queries associated with images that lack strongly related visual cues.
    \item \textit{Challenges in Recognizing Similar Objects (Visual Illusion)}. In categories with a broad range of sub-classes such as animals, plants, and food, GPT-4V demonstrates a limited capacity to distinguish between similar items. This limitation results in a notably lower accuracy in answering knowledge-intensive questions within these domains, which may be attributed to visual illusion and LLMs hallucination~\cite{zhang2023siren}.
    \item \textit{Inadequate Integration of Visual and Knowledge Dimensions}. The dataset predominantly targets single-hop visual knowledge questions and answers. The factual inaccuracies observed across all figures suggest a weak integration of visual recognition with relevant knowledge. For instance, the last example in Figure~\ref{fig:infoseek_2} shows that while GPT-4V can identify a well-known bridge, it fails to associate it with the correct contextual knowledge.
    \item \textit{Overreliance on Visual Clues, Overlooking Textual Queries}. There are instances where GPT-4V heavily bases its responses on visual elements, neglecting the textual aspect of the questions. An example of this is the third case in Figure~\ref{fig:infoseek_2}, where the model focuses on the Blackfriars Bridge in the image, ignoring the actual question about the preceding river crossing upstream. Similarly, the first case in Figure~\ref{fig:infoseek_5} also exemplifies this issue.
    
\end{itemize}

\subsection{Conclusion}
In this section, we mainly explore the performance of GPT-4V and open-source MLMs on fine-grained world knowledge-intensive VQA. Our experimental findings and case studies reveal that while all MLMs struggle to answer questions that require entity-specific knowledge, GPT-4V demonstrates a notable superiority over open-source MLMs in these scenarios. However, we also observe that \textit{GPT-4V tends to be overly cautious, often opting not to respond to queries associated with images that lack strongly related visual cues}.
A key limitation identified in these models, particularly evident in their poor performance, is their excessive dependency on visual content for understanding. This reliance becomes a significant drawback when the images do not contain information pertinent to the question, leading to a marked increase in errors. To enhance the capabilities of MLMs in fine-grained visual object knowledge tasks, further research is necessary, particularly in improving the integration and correlation of detailed visual data with contextual knowledge.

\section{Comprehensive Knowledge with Decision-Making Rationale}

Multimodal Language Models (MLMs) typically amalgamate the capabilities of pretrained large language models with visual encoders, harnessing internally stored extensive knowledge repositories. This section delves into an exploration of the reasoning and interpretation ability inherent in contemporary MLMs. Here, we present the decision-making rationales to assess the adeptness of MLMs in generating pertinent facts and logical reasoning sequences that support their inference.

\subsection{Dataset}

To achieve our evaluative purposes, the entire validation set of the A-OKVQA dataset is employed. The Augmented Open Knowledge-based Visual Question Answering (A-OKVQA~\cite{AOKVQA}) benchmark represents an evolutionary progression from its predecessor, OK-VQA. It boasts a compendium of 25,000 questions that span a wide spectrum of commonsense and world knowledge, necessitating answers. The questions curated within A-OKVQA are designed to be intellectually stimulating and conceptually varied, often demanding knowledge extrapolation beyond the presented imagery. To streamline the interaction with expansive knowledge sources, the training subset of A-OKVQA furnishes each question with corresponding rationales. These rationales comprise essential facts and logical deductions instrumental for the derivation of accurate answers.

\subsection{Results and Analysis}
\begin{figure}[t]
    \centering
    \includegraphics[width=0.98\textwidth]{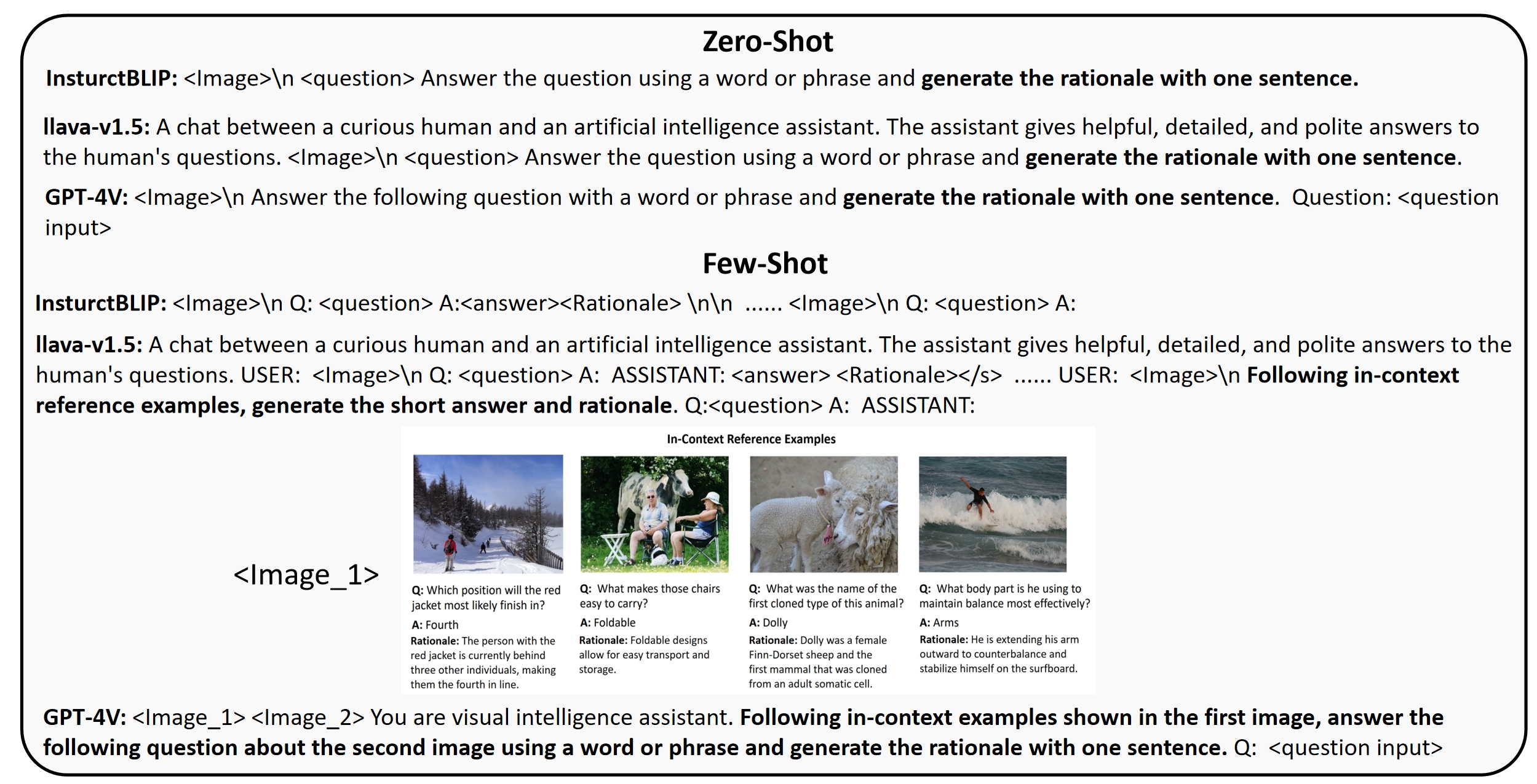}
    \caption{The illustration of Our prompting method for A-OKVQA with generating the rationale.}
    \label{fig:prompt_a_okvqa}
\end{figure}

\textbf{Prompt Method}. The zero-shot and four-shot prompting settings are depicted in Figure~\ref{fig:prompt_a_okvqa}. The current multi-modal prompting methods approach primarily focuses on generating rationales for the decision-making process, a significant shift from earlier prompting methods in Figures~\ref{fig:info-vqa-prompt} and \ref{fig:ok-vqa-prompt}.

\begin{table}[t]
\renewcommand\arraystretch{1.3}
\tabcolsep=0.12cm
\centering
\scriptsize
\caption{Benchmark results on A-OKVQA with multiple knowledge types and decision-making rationale generation. ${\clubsuit}$ and ${\dagger}$ refer to that the corresponding model is trained using the training set of OK-VQA and A-OKVQA, respectively. \textit{In addition to GPT-4V, other MLMs struggle with generating decision-making rationales under a zero-shot setting.} Len\_R indicates the average length of generated rationales. }
\label{tab:aokvqa_results}
\begin{tabular}{l|c|ccccccccc}
\toprule
Model & Answer Accuracy & BLUE-1 & BLUE-2 & BLUE-3 & BLUE-4 & Avg.BLUE & CIDER & METEOR & Len\_R \\
\hline
MiniGPT-4 (Vicuna-7b) & 36.06 &- & - & - & - & - & - & - & -\\
MiniGPT-4 (V-4-shot) & 44.0 & 26.05& 6.66 & 1.99 & 0.64 & 3.86 & 7.09 & 12.05 & 8.92\\
BLIP-2 (FlanT5-xxl) & 53.7 &- & - & - & - & - & - & - & -\\
BLIP-2 (Flan-4-shot) & 40.70 & 39.07 & 14.16 & 5.91 & 2.55 & 9.56 & 23.17 & 24.78 & 10.59\\
InstructBLIP$^{\clubsuit \dagger}$ (Vicuna-13b)& 58.30 &- & - & - & - & - & - & - & -\\
InstructBLIP$^{\clubsuit \dagger}$ (V-4-shot) & 47.77 & 41.92 & 15.06 & 6.32 & 2.81 & 10.06 & 27.06 & 24.97 & 8.62\\
InstructBLIP$^{\clubsuit \dagger}$ (FlanT5-xxl) & 62.0 &- & - & - & - & - & - & -& - \\
InstructBLIP$^{\clubsuit \dagger}$ (Flan-4-shot) & 45.67 & 41.73 & 13.95 & 5.68 & 2.17 & 9.22 & 26.10 & 27.36 & 9.84 \\
llava-v1.5-13b$^{\clubsuit \dagger}$ (0-shot) & 73.54 & - & - & - & - & - & - & - & -\\
llava-v1.5-13b$^{\clubsuit \dagger}$ (4-shot) & 62.71 & 36.80 & 12.45 & 5.0 & 1.97 & 8.20 & 24.97 & 29.50 & 15.70\\
\hline
Gemini (0-shot) & 70.48 & 35.76 & 11.67 & 4.49 & 1.61 & 7.41 & 24.33 & 31.84 & 16.88 \\
GPT-4V (0-shot) & 73.0 & 28.99 & 7.43 & 2.22 & 0.77 & 4.39 & 10.25 & 29.68 & 22.90\\
GPT-4V (4-shot) & \textbf{74.67} & 33.72 & 9.89 & 3.41 & 1.20 & 6.08 & 17.25 &32.20 & 19.18 \\
\bottomrule
\end{tabular}
\end{table}

\textbf{Comparison and Analysis}.
Based on the experimental data presented in Table~\ref{tab:aokvqa_results}, it is evident that GPT-4V (4-shot) outperforms other models, including Gemini, in terms of performance. Unlike GPT-4V, other MLMs cannot generate rationale unless they are provided with in-context examples that include rationale as a reference. Consequently, we have excluded the rationale evaluation results for these MLLMs from our presentation.
It is noteworthy that the accuracy of MLLMs, which have been trained on OK-VQA and VQA datasets such as InstructBLIP and Llava-v1.5-13b, generally declines in a few-shot setting (with rationale). GPT-4V is the exception, maintaining or improving accuracy under these conditions. This trend reveals significant limitations in the in-context learning abilities of current MLLMs.
As shown in Figure~\ref{fig:prompt_a_okvqa}, we introduced in-context examples to GPT-4V using image inputs, and performance improvement further substantiated its effectiveness.
In terms of rationale generation, automatic evaluation metrics such as BLEU and CIDEr suggest that open-source MLLMs perform better. This superior performance may be linked to their tendency to generate more concise rationales. In contrast, GPT-4V typically produces more detailed and extensive rationales in response to human queries, which could explain its different performance on these metrics.

\begin{table}[t]
\renewcommand\arraystretch{1.3}
\tabcolsep=0.10cm
\centering
\footnotesize
\caption{Human evaluation on a subset for generated answers and rationales. }
\label{tab:aokvqa_human_results}
\begin{tabular}{l|c|ccc}
\hline
Model & Answer Accuracy $\uparrow$ & Consistency$\uparrow$ & Sufficiency$\uparrow$ &  Factual Correctness$\uparrow$\\
\hline
InstructBLIP$^{\clubsuit \dagger}$ (Flan-4-shot) & 43 & 2.59 & 2.25 & 2.01\\
InstructBLIP$^{\clubsuit \dagger}$ (V-4-shot) & 51 & 1.93 & 1.42 & 1.68\\
llava-v1.5-13b$^{\clubsuit \dagger}$ (4-shot) & 65 & 3.4 & 3.33 & 3.23\\
GPT-4V (4-shot) & \textbf{72} & 3.77 & 4.04 & 3.81\\
\hline
\end{tabular}
\end{table}
\begin{figure}
    \centering
    \includegraphics[width=0.98\textwidth]{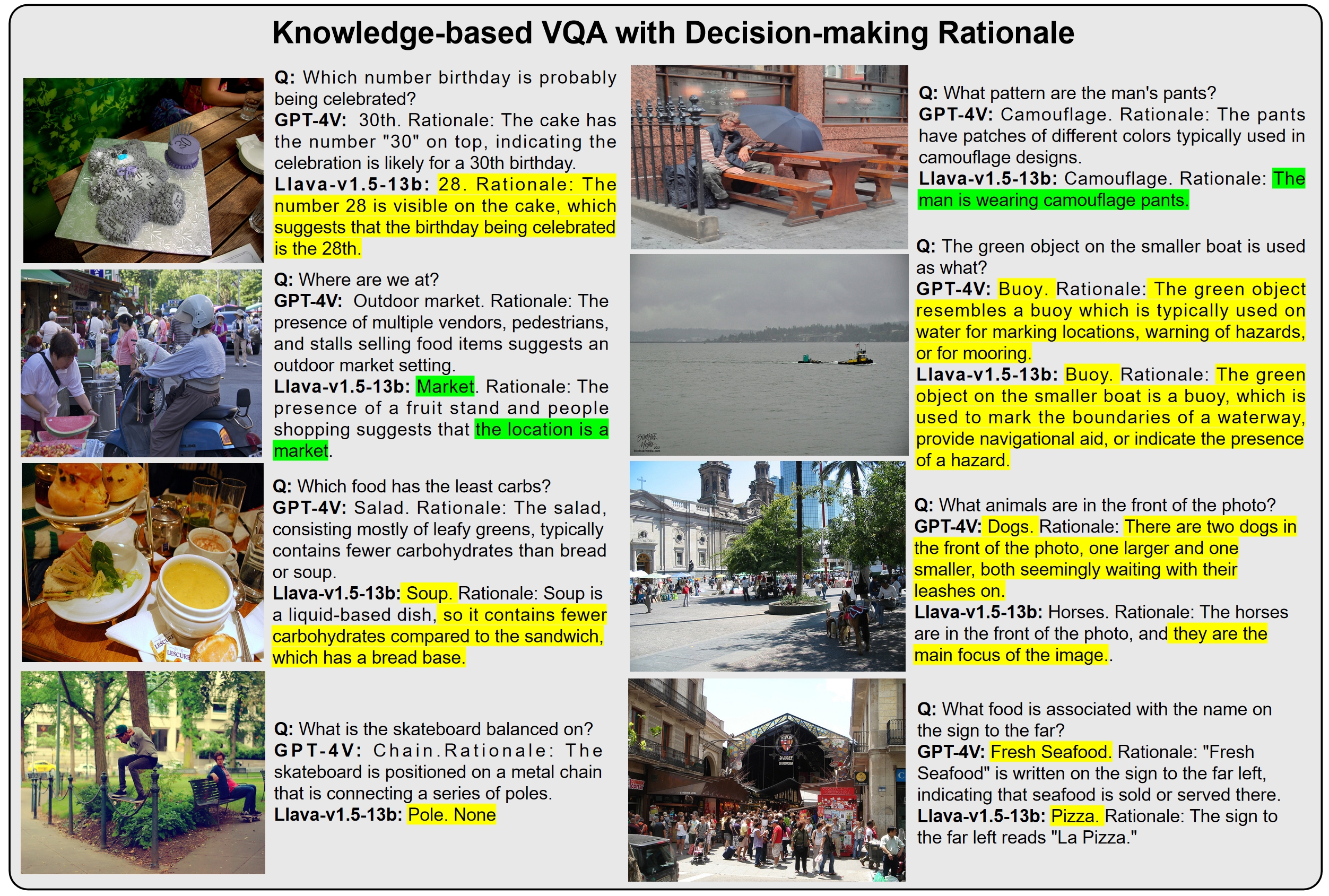}
    \caption{Eight cases generated by GPT-4V and Llava-v1.5-13b with four-shot prompt method. Words with a \textcolor[RGB]{255,216,0}{yellow} and \textcolor{green}{green} undertones indicate incorrect and inadequate responses, respectively.}
    \label{fig:cases_a_okvqa}
\end{figure}

\textbf{Human Evaluation}. 
Automatic evaluation metrics often fail to accurately assess the quality of generated content. Thus, we incorporate human evaluation to comprehensively assess the generated content of four multi-modal large models. The evaluation primarily focuses on the accuracy of the generated answers and the quality of the rationale. We first randomly selected 100 examples from the test set and then employed four annotators to score the four responses for each example. Each example was blindly reviewed and evaluated on four dimensions: Answer Accuracy (scoring 0 or 1), Consistency (range [0, 3, 5]), Sufficiency (range [0, 3, 5]), and Factual Correctness (range [0, 3, 5]). The final statistical results are presented in Table~\ref{tab:aokvqa_human_results}.
It is evident that GPT-4V excels in answer accuracy and decision-making rationale quality, significantly outperforming other open-source MLMs. Although the rationale score is marginally lower in automatic evaluation indicators, from a manual evaluation standpoint, the quality of the rationale generated by GPT-4V is still best.

\textbf{Case Study}. Figure~\ref{fig:cases_a_okvqa} presents a comparative display of samples produced by GPT-4V and Llava-v1.5. Upon evaluating the rationales provided by these samples, along with the accuracy of their responses, it is apparent that GPT-4V outperforms Llava-v1.5. Specifically, the quality of the rationale tied to the generative process reveals that GPT-4V's output is more detailed and provides a richer quantity of information. In contrast, the rationales generated by Llava-v1.5 tend to be somewhat vague and occasionally suffer from hallucinations, particularly when elaborating on visual elements. This deficiency is exemplified in the penultimate example on the right, where Llava-v1.5 describes the image's main focus as dogs. It's inconsistent with the image content.
Further analysis of incorrect responses sheds light on a recurring issue: GPT-4V and Llava-v1.5 occasionally deliver inaccurate answers mainly due to a misunderstanding of the visual content within the images. This tendency is particularly pronounced in error samples from Llava-v1.5, suggesting room for improvement in its visual information processing and instruction understanding capabilities.

\subsection{Conclusion}
In this section, we mainly evaluate the answer accuracy and decision-making rationale generation quality of GPT-4V and open-source MLMs. The evaluation results demonstrate that open-source MLMs can achieve competitive performance comparable to GPT-4V in terms of answer accuracy, and this is primarily due to their utilization of relevant data sets during the multimodal instruction-tuning stage. Nonetheless, these models struggle to generate rationales based on instructions without prior contextual references, limiting their ability to comprehend various instructions.  In addition, under a few-shot setting, the quality of the rationales generated by open-source MLMs was found to be less than GPT-4V. While these MLMs are capable of generating decision-making rationales, their answer accuracy drops drastically. In short, our findings reveal that the visual understanding ability of current open-source MLMs lags significantly behind that of GPT-4V, and their in-context learning ability requires further enhancement. Significantly, the few-shot setting way for GPT-4V improves the overall performance of answer accuracy and the quality of generated rationales.

\section{Improving MLMs with Visual Knowledge Enhanced Approaches}
\begin{figure}[t]
    \centering
    \includegraphics[width=0.98\textwidth]{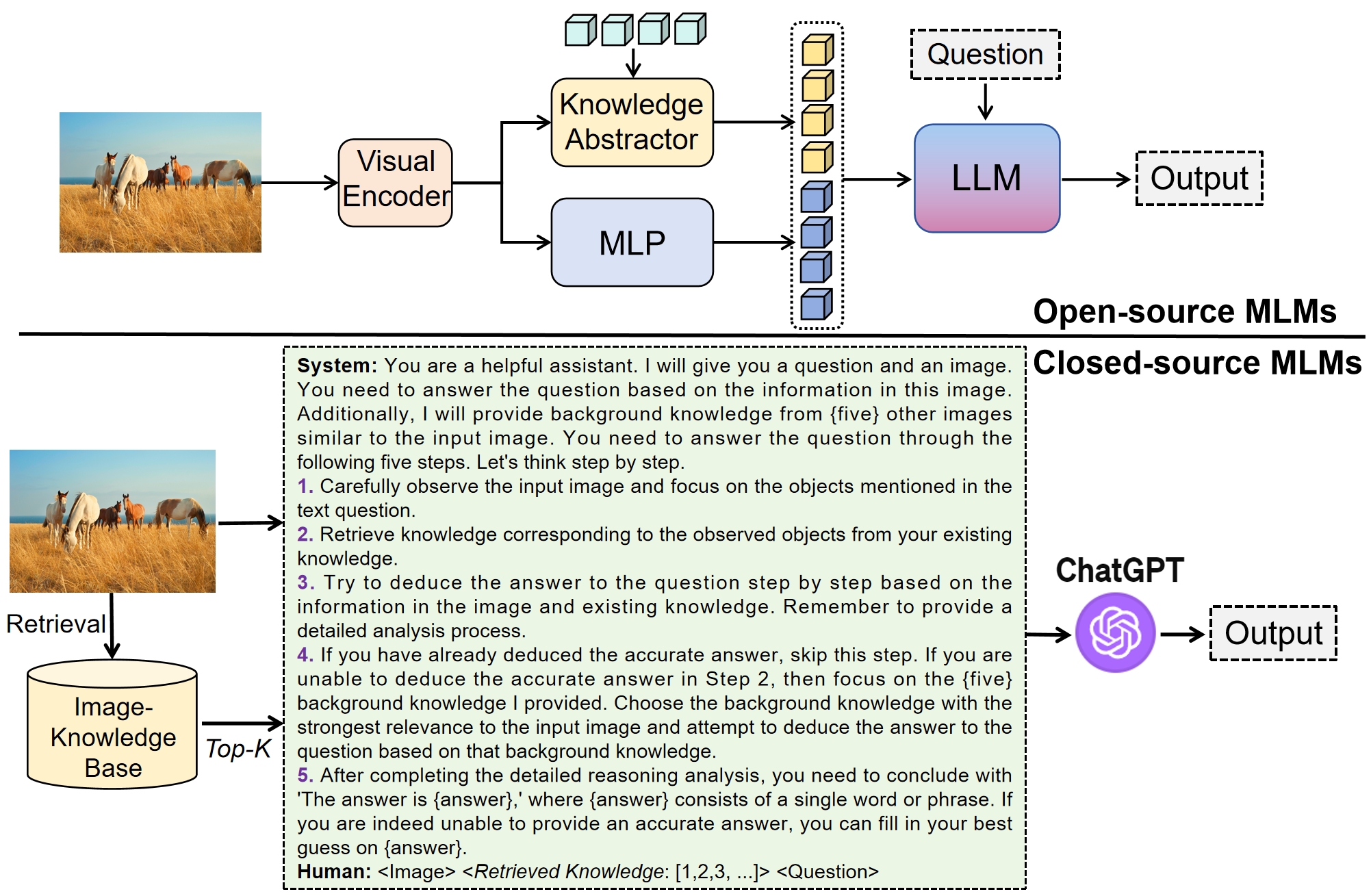}
    \caption{An overview of visual knowledge enhanced training strategy and multimodal retrieval-augmented generation. }
    \label{fig:knowledge_models}
\end{figure}

\begin{figure}[t]
    \centering
    \includegraphics[width=0.92\textwidth]{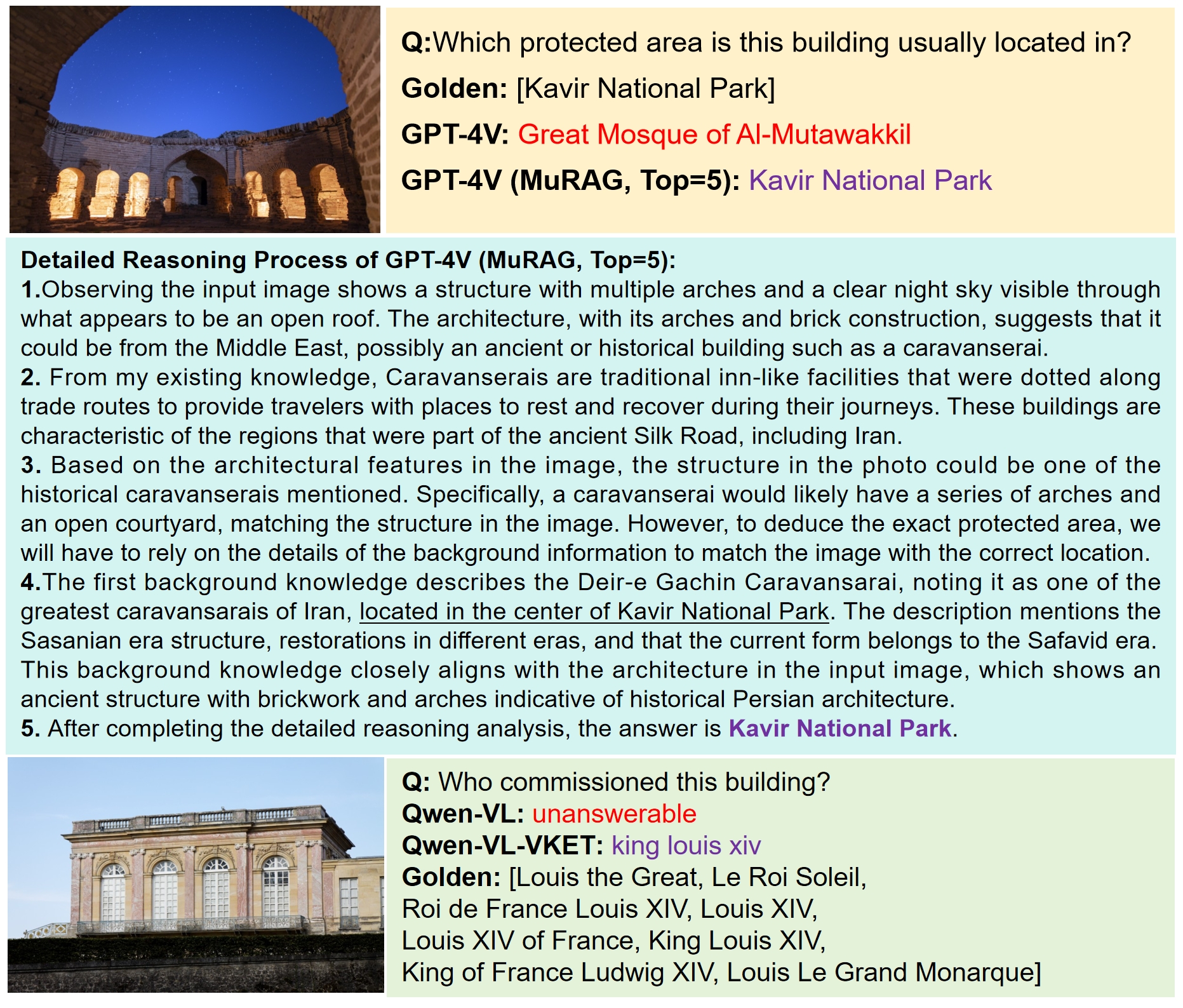}
    \caption{An illustration of cases generated by GPT-4V and Qwen-VL. Red and purple words indicate the incorrect and correct answers, respectively.}
    \label{fig:case_models}
\end{figure}

\begin{table}[t]
\renewcommand\arraystretch{1.10}
\centering
\small
\caption{Ablation study about visual knowledge enhancement approaches. ``VKET'' refers to the visual knowledge-enhanced training strategy. ``MuRAG'' represents the multimodal retrieval-augmented generation method. and ``*'' indicates that we remove the designed reasoning steps and directly splice the retrieved knowledge with the question and image. }
\scalebox{1.0}{
\begin{tabular}{l|ccc}
\hline
Model &  Commonsense $\uparrow$ & Fine-grained Knowledge$\uparrow$\\
\hline
Qwen-VL-7b$^{\clubsuit \dagger}$ (0-shot) & 70.30 & 16.24 \\
+ VKET & \textbf{71.37} & \textbf{17.12} \\
llava-v1.5-7b$^{\clubsuit \dagger}$ (0-shot) & 60.33 & 10.67 \\
+ VKET & 68.96 & 12.61\\
\hline
GPT-4V (0-shot) & 64.28 & 26.62\\
GPT-4v (Zero-COT)  & 59.33 & 24.91\\
GPT-4V (MuRAG, Top=3) & 65.22 & 27.33 \\
GPT-4V (MuRAG, Top=5) & \textbf{66.82} & \textbf{29.03}\\
GPT-4V (MuRAG$^*$, Top=5) & 63.41 & 26.55 \\
GPT-4V (MuRAG, Top=8) & 62.81 & 26.91\\
\hline
\end{tabular}}

\label{tab:knowledge_enhancement}
\end{table}

In this section, we explore how to improve the performance of MLMs on knowledge-intensive VQA, highlighting the future need for the advancement of MLMs. We introduce two simple and effective visual knowledge-enhanced approaches:
\begin{itemize}[leftmargin=*]
    \item \textbf{Visual Knowledge Enhanced Training}. In the pretraining phase of open-source MLMs, there is a predominant focus on image-text captioning data~\cite{gao2023llama,li2023lmeye,liu2023improved,Qwen-VL}, with less emphasis on the visual knowledge dimension. Based on this finding, we propose incorporating visual knowledge data during the pre-training phase to enhance the visual representation of an image with the visual knowledge representation. Our approach begins with sourcing images and their contextual descriptions from WiT's Wikipedia~\footnote{https://github.com/google-research-datasets/wit}. As these descriptions typically provide background knowledge associated with the images, they serve as suitable visual knowledge pre-training data. We collect about 2 million image-knowledge pairs\footnote{We release these image-knowledge pairs: \url{https://github.com/HITsz-TMG/Cognitive-Visual-Language-Mapper.}} and then employ a four-layer Q-former~\cite{li2023blip2} as a Visual Knowledge Abstractor to gain the knowledge-dimension representation of an image, with a fixed length of 32. This extracted visual knowledge representation is combined with the original visual representation and we fed it into the LLM, providing a comprehensive representation of the image. The detailed training details are shown in the Appendix \textcolor{red}{E}. 

    \item \textbf{Multimodal Retrieval-Augmented Generation}. As we know, retrieval-augmented generation (RAG)~\cite{chen2022murag, lewis2020retrieval,zhao2023retrieving} allows LLMs to access a vast array of external knowledge sources, enabling them to better understand the context of a given task and generate more accurate and relevant responses. Hence, we use the above image-knowledge pairs as the knowledge base and retrieve the corresponding visual knowledge based on the image similarity, which is calculated by the pretrained visual encoder of CLIP~\cite{radford2021learning}. The specific knowledge-based reasoning way is presented in Figure~\ref{fig:knowledge_models}.
\end{itemize}

\textbf{Comparison and Analysis}. The enhanced performance of models using visual knowledge-enhanced approaches is presented in Table~\ref{tab:knowledge_enhancement}. It is evident that both the visual knowledge-enhanced training strategy and the multimodal knowledge retrieval augmented method improve MLMs' performances in VQA tasks that require common sense and detailed world knowledge. Furthermore, a comparison of the volume of retrieved knowledge reveals that the precision of this knowledge critically influences reasoning accuracy. Additionally, our multi-step reasoning approach, devised to leverage visual information and knowledge, effectively mitigates the effects of noise in the retrieved knowledge. We also introduce two cases in Figure~\ref{fig:case_models} to present the effectiveness and reasoning process of two visual knowledge-enhanced approaches.

\section{Conclusion and Future Work}

This study primarily focuses on assessing the performance of multimodal large models, with a specific emphasis on GPT-4V, across diverse knowledge-intensive VQA tasks. The key conclusions and discussions drawn from our evaluation are outlined below:
\begin{itemize}
    \item \textbf{Long-Tail Knowledge Reasoning poses challenges for Multi-Modal Large Models}. Our analysis, as detailed in Tables~\ref{tab:okvqa_results} and \ref{tab:infoseek_results}, highlights a significant variance in the reasoning capabilities of multimodal large models (MLMs) across diverse knowledge domains, encompassing both commonsense and fine-grained world knowledge. This variance is most pronounced in scenarios that involve human-challenging entity knowledge, where MLMs struggle noticeably. A potential explanation for this discrepancy lies in the imbalanced distribution of data used during the models' training phases. This imbalance seems to disproportionately favor certain types of knowledge over others, leading to a disparity in the models' adeptness at handling a wide range of knowledge-intensive tasks. Addressing this imbalance is crucial for enhancing the overall reasoning proficiency of MLMs.

    \item \textbf{GPT-4V and other MLMs show performance limitations in fine-grained world knowledge Q\&A.}. Our investigations reveal that GPT-4V, along with other MLMs, exhibits notable deficiencies in fine-grained world knowledge question-and-answer scenarios. Specifically, GPT-4V encounters several distinct challenges: Reluctance to Respond Due to Insufficient Context; Challenges in Recognizing Similar Objects (Visual Illusion); Inadequate Integration of Visual and Knowledge Dimensions; Overreliance on Visual Clues, and Overlooking Textual Queries. More detailed descriptions are presented in Section~\ref{fine_grained}. Understanding and addressing these shortcomings is vital for improving the efficacy of MLMs in complex question-and-answer tasks that require a nuanced understanding of both visual and textual information.

    \item \textbf{Visual understanding should integrate comprehensive knowledge during visual-language training}. In our evaluation of knowledge-intensive visual question answering, which primarily involves single-hop reasoning, we observe a notable limitation in current MLMs. These models, when presented with questions that require understanding the background knowledge of objects depicted in images, frequently fail to provide accurate answers. Our observations, as illustrated in Figures 6-10, indicate that while powerful GPT-4V can often correctly identify the contents of an image, it falls short in fully understanding the broader context and significance of these visual elements. This gap in comprehension often leads to either refusal to respond or incorrect answers. To mitigate this issue, we suggest an expansion of visual-language training to encompass a wider range of knowledge about visual objects. By integrating more extensive background knowledge into the training process, we aim to improve the models' ability to comprehend visual content and accurately respond to complex and knowledge-intensive visual questions. Such an enhancement would not only increase the accuracy of responses but also contribute to a more nuanced understanding of visual content.

    \item \textbf{GPT-4V effectively utilizes composite image-based in-context learning}. The superior image comprehension of GPT-4V, especially in processing images with multiple sub-images, sets it apart from other open-source MLMs. Our approach involved feeding contextual references into GPT-4V via a composite image. This technique, as evidenced by our experimental results (referenced in Table~\ref{tab:aokvqa_results}), indeed enables GPT-4V to generate more accurate answers and rationales. Importantly, it reduces the token input requirement, thereby increasing the inference efficiency of large models. However, it's crucial to note that the effectiveness of this method is heavily dependent on the inherent image understanding capabilities of models. Open-source MLMs do not excel in understanding composite images that contain rich information.

\end{itemize}

\bibliographystyle{ieee_fullname}
\bibliography{reference}

\end{document}